\documentclass[11pt]{article}

\usepackage[preprint]{acl}

\usepackage{times}
\usepackage{latexsym}

\usepackage[T1]{fontenc}

\usepackage[utf8]{inputenc}

\usepackage{microtype}

\usepackage{inconsolata}

\usepackage{graphicx}

\usepackage{tikz}
\usetikzlibrary{positioning}

\usepackage{booktabs}
\usepackage{threeparttable}
\usepackage{soul}
\usepackage{amsmath}
\usepackage{amssymb}
\usepackage{multirow}
\usepackage{array}
\usepackage{acronym}
\usepackage[inline]{enumitem} 
\setlist[itemize]{leftmargin=*, nosep}
\setlist[enumerate]{leftmargin=*,nosep,label=(\roman*)}

\newcommand{\minipm}[1]{{\scriptsize$\pm$#1}}

\newcommand{\rotgroup}[2]{%
  \multirow{#1}{*}{%
    \rotatebox{90}{\parbox{1.5cm}{\centering #2}}
  }%
}

\acrodef{PIE}{potential idiomatic expression}
\acrodef{LLM}{large language model}
\acrodef{MWE}{multi-word expression}
\acrodef{ISP}{idiomatic sentence paraphrasing}

\title{IdioLink: Retrieving Meaning Beyond Words\\ Across Idiomatic and Literal Expressions}

\author{
 \textbf{Kai Golan Hashiloni\textsuperscript{1,2}},
 \textbf{Daniel Fadlon\textsuperscript{1,2}},
 \textbf{Lior Livyatan\textsuperscript{1,2}},
 \textbf{Ofri Hefetz\textsuperscript{1,2}},
\\
 \textbf{Jiahuan Pei\textsuperscript{3}},
 \textbf{Kfir Bar\textsuperscript{1,2}},
\\
 \textsuperscript{1}Data Science Institute, Reichman University, Herzliya, Israel,
\\
 \textsuperscript{2}Efi Arazi School of Computer Science, Reichman University, Herzliya, Israel,
\\
 \textsuperscript{3}Vrije Universiteit Amsterdam, The Netherlands
\\
\small
   \texttt{\href{mailto:kai.golanhashiloni@post.runi.ac.il}{kai.golanhashiloni@post.runi.ac.il}}, 
   \texttt{\{daniel.fadlon, lior.livyatan, ofri.hefetz\}@post.runi.ac.il}, 
\\
\small
   \texttt{j.pei2@vu.nl}, 
   \texttt{kfir.bar@runi.ac.il}
}

\begin{document}
\maketitle
\begin{abstract}
Idioms pose a fundamental challenge for language models, as their meaning cannot be inferred from surface form alone.
Understanding such expressions, therefore, requires semantic abstraction beyond lexical overlap.
We introduce \textbf{IdioLink}, a retrieval benchmark designed to test whether models can link idiomatic expressions to conceptually equivalent meanings expressed in literal or paraphrased forms. 
IdioLink comprises 10,700 documents and 2,140 queries, spanning 107 idioms with both literal and figurative uses.
Each document and query is annotated with spans that convey the core meaning.
Evaluating strong embedding baselines (e.g., BGE, E5, Contriever, and Qwen), we show that current models struggle to retrieve equivalent meanings across divergent surface realizations, relying instead on topical and shallow semantic cues. IdioLink exposes key gaps in idiom-aware semantic retrieval and provides a challenging testbed for future models.

\end{abstract}

\section{Introduction}

Understanding language goes beyond mapping words to meanings: it involves constructing conceptual interpretations that often depart from surface form.
Figurative language, including idioms, metaphors, and other conventionalized expressions, illustrates this challenge.
\Acfp{PIE} like ``break the ice'' or ``spill the beans'' convey meanings that cannot be compositionally derived from their constituent words, yet are effortlessly understood by humans based on the context they appear in, which determines whether the expression is intended figuratively or literally.
Decades of work in linguistics and cognitive science show these expressions are central to human communication and drawing on rich conceptual, cultural, and contextual knowledge \citep{pollio1977psychology,weinreich1969problems,jackendoff1997architecture}, making them a crucial benchmark for evaluating semantic understanding beyond words.

\begin{figure}[t]
    \centering
    \includegraphics[scale=0.25]{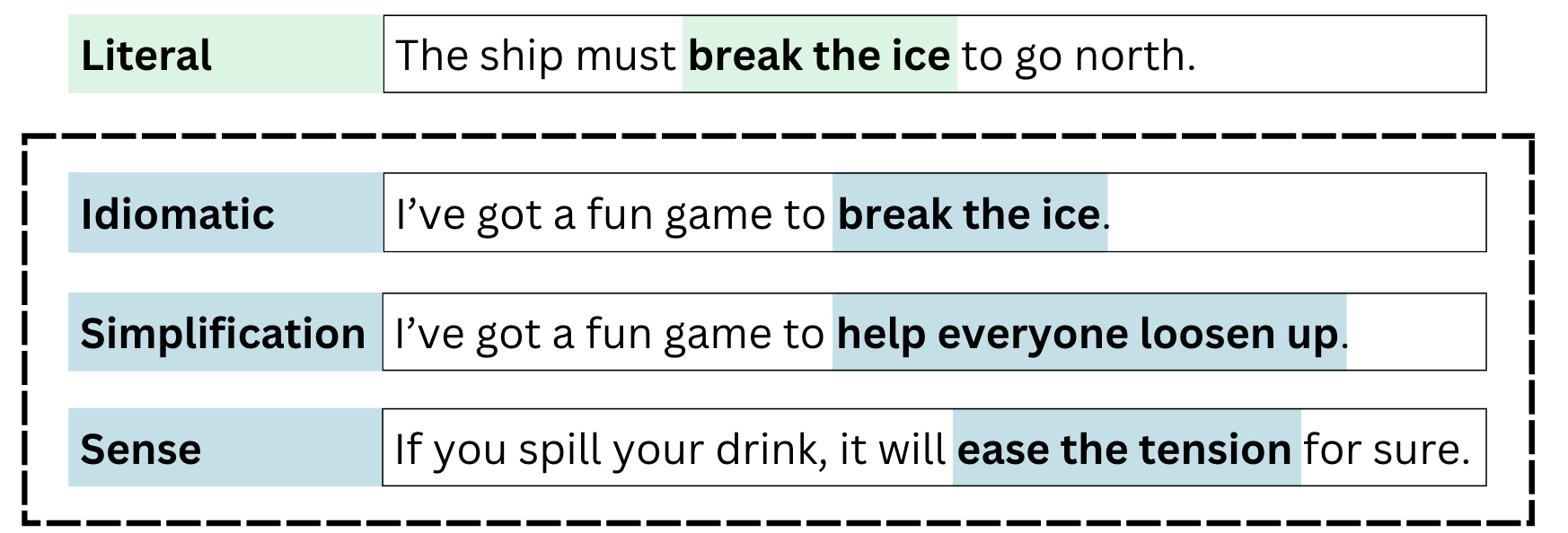}
    \caption{The four different document types for a given \acl{PIE} in IdioLink.}
    \label{fig:sample}
\end{figure}

Recent advances in \acp{LLM} have substantially improved performance on modeling figurative language, including idiom identification \citep{tedeschi-etal-2022-id10m,hashiloni-etal-2025-easy} and figurative-literal expression disambiguation \citep{arslan-etal-2025-using,mi-etal-2025-rolling}.
However, it remains unclear whether LLMs can reliably align expressions with the same meanings across figurative, literal, or paraphrased forms.
Embedding and contrastive models often rely on lexical overlap rather than deeper semantic equivalence~\citep{wang2022sncsecontrastivelearningunsupervised,Miao_2023,shi-etal-2023-osscse}, causing retrieval failures when query meaning is expressed figuratively or paraphrased, e.g., ``break the ice'' vs. ``ease the tension''.
This gap is especially problematic in real-world scenarios where users encounter unfamiliar idioms and seek semantically equivalent expressions to understand their usage.

In this work, we introduce \textbf{IdioLink}, a benchmark designed to evaluate concept-level retrieval across \texttt{idiomatic}, \texttt{literal}, \texttt{simplification}, and \texttt{sense} document types, as shown in Figure~\ref{fig:sample}.
Each instance is annotated with a marked span representing the core semantic unit of a \ac{PIE} and paired with documents containing a span expressing the same meaning from different types.
We formulate a retrieval task in which, given a query containing a marked span with a \ac{PIE}, used either figuratively or literally, the model retrieves all documents whose marked spans convey the same meaning, without access to span annotations in the documents.
To the best of our knowledge, this is the first setup that enables evaluation of whether models retrieve documents based on shared conceptual meaning, whether expressed figuratively or literally, rather than superficial lexical or topical similarity.

We evaluate 24 embedding models and their variants, comparing zero-shot and fine-tuned settings, encoding queries w/ or w/o instructions, and employing standard \textit{sentence} or \textit{span} embeddings methods.
Our results reveal that even state-of-the-art embedding models struggle to consistently retrieve semantically equivalent expressions when form diverges from meaning, relying heavily on lexical overlap and topic similarity. 
These findings highlight a persistent gap between surface-level semantic matching and genuine conceptual understanding.

In summary, our contributions are as follows:
\begin{itemize}
    \item We introduce a new retrieval task centered on idiomatic expressions, enabling controlled analysis of conceptual similarity beyond surface form.
    \item We release IdioLink, a new benchmark for evaluating concept-level retrieval across figurative, literal, and paraphrased expressions.
    \item We provide an extensive evaluation of modern embedding models, revealing systematic limitations in their ability to represent and retrieve shared meaning.
\end{itemize}


The IdioLink dataset, prompts, and evaluation code are publicly released to ensure transparency and reproducibility under the CC-BY-4.0 (dataset) and Apache-2.0 (code) licenses: \url{https://github.com/Intellexus-DSI/IdioLink}.



\section{Related Work}

\subsection{Idioms as a Core Challenge in NLP}

\Acp{MWE} pose a longstanding challenge in NLP due to their syntactic variability and non-compositional semantics \citep{constant-etal-2017-survey}. 
Earlier approaches relied on rule-based, statistical, and distributional methods \citep{cook-etal-2007-pulling, fazly-etal-2009-unsupervised, gharbieh-etal-2016-word}, while recent work predominantly uses Transformer-based architectures for MWE identification and interpretation \citep{taslimipoor-etal-2020-mtlb, zeng-bhat-2021-idiomatic, savary-etal-2023-parseme}.


Idioms form a particularly challenging subclass of MWEs because their meanings cannot be compositionally inferred from their constituent words \citep{mwe_nlp_baldwin}. 
In NLP, idiom \emph{identification} focuses on detecting idiomatic spans in context, whereas idiom \emph{classification} assumes candidate expressions are pre-specified. 
A wide range of datasets has been proposed for these tasks, including VNC-Tokens \citep{cook2008vnctokens}, IDIX \citep{sporleder-etal-2010-idioms}, SemEval benchmarks \citep{korkontzelos-etal-2013-semeval, tayyar-madabushi-etal-2022-semeval}, and larger modern resources such as MAGPIE \citep{haagsma-etal-2020-magpie} and EPIe \citep{saxena2020epiedatasetcorpuspossible}.



While our work focuses on English, several datasets extend idiom and MWE evaluation to multilingual settings. 
PARSEME \citep{ramisch-etal-2020-edition, savary-etal-2023-parseme} and AlphaMWE \citep{han-etal-2020-alphamwe} provide broad multilingual MWE coverage beyond idioms, with PARSEME 2.0 \citep{scholivet-etal-2026-edition} additionally supporting paraphrasing tasks. 
Other multilingual resources include ID10M \citep{tedeschi-etal-2022-id10m}, Dodiom \citep{Eryi_it_2022}, and MIDAS \citep{kim-etal-2025-memorization}. 
Recent work further explores figurative-language understanding, idiom disambiguation, and cross-lingual semantic retrieval \citep{park-etal-2025-fluid, mi-etal-2025-rolling, aslantas-gungor-2026-unified}.

Neural approaches have been applied to both idiom classification and identification \citep{BRISKILAL2022102756,he-etal-2024-enhancing}. 
More recently, LLMs have achieved strong performance through prompt-based inference, often matching or surpassing fine-tuned systems \citep{hashiloni-etal-2025-easy, de-luca-fornaciari-etal-2024-hard,mi-etal-2025-rolling,phelps-etal-2024-sign}. 
LLMs have also been used to support the construction of idiom datasets and corpora \citep{arslan-etal-2025-using}.

Another related direction is \acf{ISP}, which focuses on transforming idiomatic expressions into literal paraphrases or vice versa. Prior work has explored unsupervised, weakly supervised, and controllable paraphrasing settings, while consistently highlighting data-scarcity challenges \citep{zhou-etal-2021-pie, qiang-etal-2023-chinese-idiom, zhou2021idiomaticexpressionparaphrasingstrong}.

However, no existing work is designed for the idiom-centered retrieval task we address.

\subsection{Text Retrieval and Embedding Models}

Text retrieval has undergone rapid evolution with the advent of neural embedding models, which map queries and documents into a shared semantic space.
Early dense retrieval approaches such as Sentence-BERT \citep{reimers-gurevych-2019-sentence} and subsequent contrastive learning frameworks \citep{gao-etal-2021-simcse, izacard2022unsuperviseddenseinformationretrieval} demonstrated strong performance by learning semantically meaningful sentence representations.
More recent models scale this paradigm through larger architectures, multilingual training, and instruction-based supervision \citep{wang2024textembeddingsweaklysupervisedcontrastive, wang2024multilinguale5textembeddings, su-etal-2023-one, muennighoff2025generativerepresentationalinstructiontuning}.

A growing body of work explores instruction-aware retrieval, where natural language instructions guide embeddings toward task-specific representations.
Models such as InstructOR \citep{su-etal-2023-one} and FollowIR \citep{weller-etal-2025-followir} show that instructions can substantially improve retrieval, motivating our use of instruction-augmented inputs to steer models toward idiomatic or literal interpretations without changing model parameters.

Another relevant direction concerns the granularity of representation and context modeling.
Standard retrieval pipelines typically embed entire documents or passages, which may obscure fine-grained semantic distinctions.
To address this, prior work has explored chunking strategies based on fixed windows, sentence boundaries, or semantic similarity \citep{lewis2021retrievalaugmentedgenerationknowledgeintensivenlp, kamradt2024fivelevels}.
More recently, methods such as late interaction \citep{santhanam-etal-2022-colbertv2, xiao-etal-2024-jina} and late Chunking \citep{günther2025latechunkingcontextualchunk} aim to preserve contextual information while enabling localized matching.
These approaches prove that retaining broader context during representation learning can substantially improve retrieval quality, especially when meaning depends on subtle contextual cues.

Our work builds on these insights while addressing a distinct challenge: retrieving semantically equivalent documents based on underlying conceptual meaning. 
Unlike prior work focused on topical relevance or general similarity, IdioLink tests whether models can distinguish literal from idiomatic meanings despite similar surface forms, situating the benchmark at the intersection of information retrieval and figurative language understanding.

\section{The IdioLink Dataset}

\subsection{Problem Formulation}
\label{ssec:problem_formulation}

We formulate IdioLink as a retrieval task, consisting of a non-overlapping index $\mathcal{I}$ and queries $\mathcal{Q}$.
Both $\mathcal{I}$ and $\mathcal{Q}$ contain similarly structured texts, each associated with a target \ac{PIE} $p$ and containing a marked span that conveys the meaning of $p$, either idiomatically or literally.
We define four types of documents (examples are in Figure~\ref{fig:sample}): 
\begin{enumerate*}
    \item \texttt{literal}, where $p$ is used in its compositional meaning;
    \item \texttt{idiomatic}, where $p$ appears in its figurative meaning;
    \item \texttt{simplification} is a copy of \texttt{idiomatic}, but replaces $p$ with an explicit explanation of its meaning that fits the context, and;
    \item \texttt{sense} expresses the same idea as the idiomatic usage of $p$, similarly to \texttt{simplification} but as a standalone, independent document.
\end{enumerate*}

We design \texttt{literal} and \texttt{idiomatic} documents to test whether models can distinguish literal and idiomatic usages \acp{PIE}. 
The \texttt{simplification} documents evaluate whether models capture the equivalence between an idiom and its literal paraphrase, while \texttt{sense} documents probe retrieval based on shared abstract meaning beyond lexical overlap.

Queries are generated exclusively as either \texttt{literal} or \texttt{idiomatic} document types.
Given a query $q \in $ \{\texttt{literal}, \texttt{idiomatic}\}, which contains a \ac{PIE} $p$, the goal is to retrieve all documents in $\mathcal{I}$ containing spans expressing the same underlying meaning of $p$ based on its usage in $q$.
If $q$ is \texttt{literal}, the relevant documents are other \texttt{literal} instances of the same $p$.
If $q$ is \texttt{idiomatic}, we expect all three other types
(a.k.a., idiomatic types) to be retrieved.

\subsection{Dataset Construction}
All models used to construct the dataset are selected based on preliminary exploratory experiments and are used with the default temperature value of 1.
All prompts are first optimized to reduce disagreement with human annotations during a calibration phase.
The prompts are released in the project repository\footnote{Given their length, we omit them from the paper.}.
The dataset construction process, including prompt design and optimization, required approximately 6,500 API calls and incurred a total cost of approximately \$60 USD.

\subsubsection{\ac{PIE} Collection}
\label{ssec:data_def}

We collect our set of \acp{PIE} from the MAGPIE corpus \citep{haagsma-etal-2020-magpie}.  
To ensure both linguistic validity and contextual diversity, we filter \acp{PIE} using three criteria.  
First, the PIE must appear at least 30 times in MAGPIE to ensure sufficient contextual variability and reflect more common usage in the language.  
Second, we require a balanced distribution between literal and idiomatic usages, constraining the proportion of idiomatic instances to lie between 25\% and 75\%. This ensures that each expression meaningfully supports both usage types and remains genuinely ambiguous and context-dependent.  
Third, we retain only PIEs annotated with maximal annotator confidence (1.0), ensuring high annotation reliability.
This yields 118 candidate \acp{PIE}.
During manual inspection, we also removed 11 PIEs whose literal usage was judged highly unnatural or artificially constructed (e.g., ``for keeps''), resulting in 107 \acp{PIE} in the final dataset.
We partition the 107 \acp{PIE} into training (22), validation (10), and test (75) splits.

\subsubsection{Data Generation and Indexing}
\label{ssec:gen_and_index}

To promote semantic diversity and reduce topical bias, we compile a set of ten domains: \textit{Politics}, \textit{Sport}, \textit{Technology}, \textit{History}, \textit{Medicine}, \textit{Culture}, \textit{Entertainment}, \textit{Food}, \textit{Business}, and \textit{Environment}.

To construct the index $\mathcal{I}$, for every pair of a \ac{PIE} and a domain $(p, d)$, we generate ten documents: four \texttt{literal} documents and two documents for each of the idiomatic types.  
As a result, each $p$ is associated with 100 documents in total---40 literal and 60 belonging to the idiomatic types.
The query set, $\mathcal{Q}$, contains one \texttt{literal} and one \texttt{idiomatic} document for each $(p, d)$ pair. 
Specifically, we use Gemini 2.5 Pro to generate $\mathcal{I}$ and Gemini 3 Flash to generate $\mathcal{Q}$.
This separation encourages novel query generation and reduces lexical and semantic overlap between queries and documents.
Each query and document consists of one or two sentences.
Both $\mathcal{I}$ and $\mathcal{Q}$ are constructed independently for each data split, ensuring no \acp{PIE} are shared across them.

\subsubsection{Human Annotation}
\label{ssec:human_annotation}

To establish a gold-standard evaluation set, we manually annotate 13 randomly sampled  \acp{PIE} from the test split.
This set is denoted as test$_\text{gold}$ (1,300 samples), and the remaining portion of the test set, 62 \acp{PIE}, as test$_\text{silver}$ (6,200 samples). 
The index and queries of test$_\text{gold}$ are manually validated and corrected when needed.

Three annotators (demographics in Appendix~\ref{apdx:annotators_info}) are provided with the documents and queries via a shared Google Sheet and instructed to follow the annotation guidelines, summarized below.
The full guidelines are released in the project's repository.

Each document is accompanied by its corresponding $(p, d)$ pair, its type, and the marked span, and is considered valid if all of the following conditions hold:
\begin{enumerate*}
    \item It is fluent, coherent, and fully interpretable to a human reader.
    \item It contains no idiomatic expressions other than possibly $p$, which must appear at most once, and its usage, idiomatic or literal, is unambiguous.
    \item The content of the document relates to $d$.
    \item The marked span correctly and minimally captures the conceptual realization of $p$ and appears in the document.
\end{enumerate*}

Annotators may mark a document as valid, correct the span, and/or revise the text.
Each query and document receives a single annotation, with ambiguous or complex cases resolved through discussion among the annotators.

Only 125 corrections or revisions were required, with most cases (66) requiring only a span correction.
There were also literal--idiomatic mismatches (21), ambiguous realizations (16), and sentence-level rewrites for improved clarity or validity (9).
Most corrections occurred in \texttt{sense} documents (67) and \texttt{literal} documents (47).

\subsubsection{Automatic and Human Validation}
\label{ssec:validation}

To scale validation while maintaining quality, we combine automatic validation with human review.

We prompt GPT-4o mini with instructions based on the annotation guidelines (\S~\ref{ssec:human_annotation}).
The model is provided with a set of documents or queries, each represented as $(p,d)$, and is tasked with determining whether each of them is valid or invalid.
We apply self-consistency \citep{wang2023selfconsistencyimproveschainthought}, running each prompt three times and assigning the final label via majority vote.

To assess the performance of the automatic validation, we compute agreement between human annotators and the automatic validation model using the index $\mathcal{I}$ of test$_\text{gold}$.
The automatic validation model achieves an accuracy of 72.08\% and a macro-averaged F1 score of 82.54\%. 
Both metrics are calculated for the binary valid/not-valid classification task.
More importantly, we observe only 42 cases in which the model labeled a document as valid, even though the annotator found it invalid, which may cause a flaw in the dataset. 
In most of those cases, the usage of the PIE was ambiguous.
This implies that the model may miss only 3.23\% of the invalid cases, which are more important to us.
When applied to all the silver-standard texts (11,280 documents and queries in total), this process flags 2,010 documents (17.81\%) as invalid.
To avoid discarding the flagged 2,010 texts, our human annotators subsequently reviewed them to validate the model's decision, finding that most cases were, in fact, validation model errors.
Only 316 (15.72\%) were confirmed by the annotators as problematic and subsequently corrected or replaced.
Overall, we obtain 3,570 gold-standard documents and queries, accounting for approximately 27.8\% of the full IdioLink dataset.
An overview of the dataset is provided in Table~\ref{tab:idiolink}.




\begin{table}[t]
  \centering
  \small
  \begin{tabular}{l >{\raggedleft\arraybackslash}p{0.75cm} >{\raggedleft\arraybackslash}p{1.1cm} >{\raggedleft\arraybackslash}p{1.7cm}}
    \toprule
    \textbf{Split} & \multicolumn{1}{c}{\textbf{\# PIE}} & \multicolumn{1}{c}{\textbf{\# Query}} & \multicolumn{1}{c}{\textbf{\# Document}} \\
    \midrule
    \textbf{Train}           & 22   & 440   & 2,200  \\
    \textbf{Validation}      & 10   & 200   & 1,000  \\
    \textbf{Test$_{silver}$} & 62   & 1,240 & 6,200  \\
    \textbf{Test$_{gold}$}   & 13   & 260   & 1,300  \\
    \textbf{Test$_{total}$}            & 75   & 1,500 & 7,500  \\
    \midrule
    \textbf{Total}           & 107  & 2,140 & 10,700 \\
    \bottomrule
  \end{tabular}
  \caption{Overview of the IdioLink Dataset. Test$_{total}$ is the sum of Test$_{gold}$ and Test$_{silver}$.}
  \label{tab:idiolink}
\end{table}

\section{IdioLink Retrieval Benchmark Models}
\label{sec:benchmarking}
We evaluate models under zero-shot and fine-tuning settings, provided with and without task-specific instruction and explore a span-aware embedding-extraction method to control how $q$ is encoded and represented.

\subsection{Retrieval Configurations}
\paragraph{Using instructions in query construction.}
We consider two approaches to submitting queries to the model: a standard approach without additional instructions and one that provides explicit instructions to guide the model.
The instruction specifies the marked span of the \ac{PIE}, and encourages the model to infer whether the \ac{PIE} is used literally or figuratively and to retrieve documents that entail conceptually equivalent meanings.
Implementation details are described in Appendix~\ref{apdx:inst}.

\begin{figure}[htb!]
    \centering
    \includegraphics[width=0.7\columnwidth]{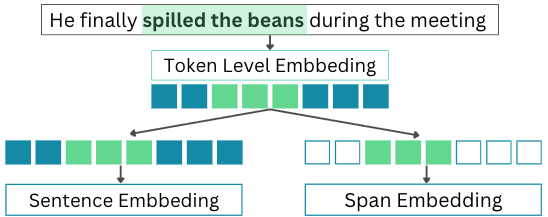}
    \caption{Sentence and span embedding. The latter employs mean-pooling over the embeddings of the span tokens only.}
    \label{fig:emb_methods}
\end{figure}

\paragraph{Using span embedding.}
We evaluate two methods for obtaining query representations.
The first is the \textit{sentence} embedding, which uses the model's default text-embedding output.
The second is \textit{span} embedding, inspired by the recent work on late chunking \citep{günther2025latechunkingcontextualchunk} and prior findings suggesting that span-level representations can better capture fine-grained semantic alignment than sentence-level similarity alone \citep{Kanerva2025}.
While standard dense retrieval typically embeds entire documents as single vectors, such representations may obscure fine-grained semantic information---particularly when the retrieval target is a localized meaning expressed as a specific span.
In contrast, late chunking first extracts token-level contextual embeddings as the model's last-layer outputs, then aggregates the embeddings of tokens corresponding to the annotated span using mean pooling.
This yields a span-level representation that remains context-aware while focusing on the segment conveying the intended meaning.
An illustration is shown in Figure~\ref{fig:emb_methods}.

\paragraph{Index and similarity calculation.} At retrieval time, queries are encoded w/ or w/o instructions and represented using sentence or span embeddings, while indexed documents are encoded without instructions as full text, reflecting a realistic retrieval setting without access to annotated spans.

\subsection{Zero-shot Setting}
\label{ssec:zero_shot_meth}

We evaluate all models in a zero-shot setting, with and without instructions, and with sentence and span embeddings, yielding four zero-shot configurations per model.
Importantly, these methods require no additional training and can be used uniformly across all architectures.

\subsection{Fine-tuning Setting}
\label{ssec:fine_tuning}

To improve retrieval of conceptual meaning beyond surface form, we fine-tune embedding models using a contrastive objective inspired by recent dense retrieval frameworks \citep{lee2024geckoversatiletextembeddings,karpukhin-etal-2020-dense}.
Our goal is to encourage alignment between semantically equivalent expressions, across literal and idiomatic realizations, while separating semantically mismatched ones.
Hyperparameters are reported in Table~\ref{tab:ft_params} (Appendix~\ref{apdx:ft_apdx}).

\paragraph{Training data.}
We construct a fine-tuning dataset $\mathcal{D}$ consisting of structured tuples:
\vspace{-4pt}
\begin{equation}
    \mathcal{D} = \{(q, p, d^{+}, \{d^{-}_i\}_{i=1}^{\alpha_N}, \{h^{-}_j\}_{j=1}^{\alpha_H})\},
\end{equation}
where each tuple contains a query $q$, its associated \ac{PIE} $p$, one positive document $d^{+}$, $\alpha_N$ ``soft'' negative documents $\{d^{-}_i\}_{i=1}^{\alpha_N}$, and $\alpha_H$ ``hard'' negative documents $\{h^{-}_j\}_{j=1}^{\alpha_H}$.

Each $q$ produces a single training tuple, so $\mathcal{D}$ contains 440 training samples, as elaborated in \S\ref{ssec:data_def}. 
A positive document $d^{+}$ contains a span that convey the same underlying meaning as $p$ in $q$.
If $q$ is of type \texttt{literal}, the corresponding $d^{+}$ must also be \texttt{literal}, that is, another document in which $p$ is used literally.
If $q$ is \texttt{idiomatic}, the corresponding $d^{+}$ is drawn from the corresponding idiomatic types, that is, a document with a span that conveys the figurative meaning of $p$.
A ``hard'' negative $h^{-}$ is a document of the same $p$, but representing the opposing interpretation:
For \texttt{literal} queries, a ``hard'' negative is a document drawn from the idiomatic types, whereas for \texttt{idiomatic} queries a hard negative is a \texttt{literal} document.
We define ``soft'' negatives $d^{-}$ as documents of any type sampled from the documents generated for other \acp{PIE}.
This construction forces each query $q_i$ to rank its corresponding positive example $d_i^+$ above both hard and soft negatives.

\paragraph{Training objective.}
We optimize the model for the following contrastive learning objective:
\vspace{-10pt}
\begin{equation}
\hspace{-1em}  
\mathcal{L}
=
-\log
\frac{
e^{\mathrm{s}(q_i, d_i^+)}
}{
\sum_{j=1}^{\alpha_N} e^{\mathrm{s}(q_i, d_{i,j}^{-})}
+
\sum_{j=1}^{\alpha_H} e^{\mathrm{s}(q_i, h_{i,j}^{-})}
}
\end{equation}
\vspace{-12pt}

where $\mathrm{s}(\cdot)$ denotes cosine similarity.
This objective encourages the model to align semantically equivalent expressions while explicitly separating misleading or contrastive realizations.

\paragraph{Configurations.}
We fine-tune a subset of models using four training configurations, aligned with those in the zero-shot setting.
During inference, queries are formulated and embedded following the same procedures used during training. 
Models trained with instructions are evaluated using both sentence-level and span-level embeddings, whereas models trained with span-level embeddings are evaluated using this method only.

\section{Experimental Settings}
\label{sec:experiments}

For details about software, hardware, running times, etc., refer to Appendix~\ref{apdx:tech_details}.

\subsection{Evaluation Metrics} \label{ssec:eval} 
As a retrieval benchmark, IdioLink is evaluated using two standard, complementary metrics (more in Appendix~\ref{apdx:metrics}):
\begin{enumerate*}
    \item nDCG@10 (binary relevance), which emphasizes ranking quality at top positions~\citep{lee2024geckoversatiletextembeddings,günther2025latechunkingcontextualchunk};
    \item R-Precision, which measures precision at rank R, the number of relevant documents per query~\citep{wu-etal-2024-kpeval,hasan-ng-2014-automatic}. 
    In IdioLink, we set R=60 for \texttt{idiomatic} queries and R=40 for \texttt{literal} queries. These values are determined by design and reflect the number of relevant documents generated for each query type during dataset construction (\S~\ref{ssec:gen_and_index}).
\end{enumerate*}
These metrics capture both early precision and overall retrieval correctness, providing a sufficient and interpretable evaluation of ranking quality.

\subsection{Models}
\label{ssec:models}
IdioLink is designed to evaluate the embedding-based semantic retrieval methods for figurative language. 
We provide an evaluation and analysis of BM25 \citep{bm25} in Appendix~\ref{apdx:bm25}.
We evaluate 24 text-embedding models, with parameter counts ranging from 100 million to 9 billion, and provide the exact checkpoint details and licenses in Table~\ref{tab:checkpoints} (Appendix~\ref{apdx:checkpoints}).
Following \citet{weller-etal-2025-followir}, we group models into two categories:

\begin{table*}[t]
  \centering
  \small
  \resizebox{\textwidth}{!}{
  \begin{tabular}{llr|rr|rr|rr|rr}
    \toprule
    &  &  & \multicolumn{4}{c}{\textbf{without instructions}} & \multicolumn{4}{c}{\textbf{with instructions}} \\
    \cmidrule(lr){4-7}
    \cmidrule(lr){8-11}
    
    &  &  & \multicolumn{2}{c}{\textbf{R-Precision}} & \multicolumn{2}{c}{\textbf{nDCG@10}} & \multicolumn{2}{c}{\textbf{R-Precision}} & \multicolumn{2}{c}{\textbf{nDCG@10}} \\
    \cmidrule(lr){4-5}
    \cmidrule(lr){6-7}
    \cmidrule(lr){8-9}
    \cmidrule(lr){10-11}
    
    & \textbf{Model} & \textbf{Param} 
    & \textbf{sentence} & \textbf{span} & \textbf{sentence} & \textbf{span} & \textbf{sentence} & \textbf{span} & \textbf{sentence} & \textbf{span} \\
    \midrule
    \rotgroup{5}{\scriptsize\texttt{No-Instruct}}
    & SBERT & 110M & 18.34 & 40.88$^{\scriptsize\blacktriangle}$ & 35.08 & 58.86$^{\scriptsize\blacktriangle}$ & 32.57 & 42.40$^{\scriptsize\blacktriangle}$ & 52.76 & 55.10$^{\scriptsize\blacktriangle}$ \\
    & Contriever & 110M & 9.60 & 38.40$^{\scriptsize\blacktriangle}$ & 21.60 & 64.80$^{\scriptsize\blacktriangle}$ & 15.10 & 42.10$^{\scriptsize\blacktriangle}$ & 30.80 & 64.10$^{\scriptsize\blacktriangle}$ \\
    & E5$_{\text{base-v2}}$ & 110M & 20.00 & 30.20$^{\scriptsize\blacktriangle}$ & 38.50 & 52.50$^{\scriptsize\blacktriangle}$ & 24.30 & 35.40$^{\scriptsize\blacktriangle}$ & 45.10 & 57.60$^{\scriptsize\blacktriangle}$ \\
    & Drama$_{\text{1B}}$ & 1B & 19.70 & 52.00$^{\scriptsize\blacktriangle}$ & 37.10 & 72.80$^{\scriptsize\blacktriangle}$ & 23.90 & 50.30$^{\scriptsize\blacktriangle}$ & 42.10 & 64.90$^{\scriptsize\blacktriangle}$ \\
    & Stella$_{\text{en\_1.5B\_v5}}$ & 2B & 18.30 & 27.60$^{\scriptsize\blacktriangle}$ & 32.00 & 41.10$^{\scriptsize\blacktriangle}$ & 18.30 & 43.60$^{\scriptsize\blacktriangle}$ & 32.00 & 58.10$^{\scriptsize\blacktriangle}$ \\
    \midrule
    \rotgroup{19}{\scriptsize\texttt{Instruction}}
    & TART-Contriever & 110M & 12.10 & 40.40$^{\scriptsize\blacktriangle}$ & 26.20 & 64.10$^{\scriptsize\blacktriangle}$ & 17.50 & 44.60$^{\scriptsize\blacktriangle}$ & 34.30 & 60.20$^{\scriptsize\blacktriangle}$ \\
    & BGE$_{\text{base-en-v1.5}}$ & 326M & 20.00 & 22.80$^{\scriptsize\blacktriangle}$ & 38.20 & 42.30$^{\scriptsize\blacktriangle}$ & 24.70 & 26.10$^{\scriptsize\blacktriangle}$ & 42.30 & 43.90$^{\scriptsize\blacktriangle}$ \\
    & InstructOR$_{\text{base}}$ & 335M & 19.80 & 52.96$^{\scriptsize\blacktriangle}$ & 38.70 & 68.37$^{\scriptsize\blacktriangle}$ & 21.30 & 47.50$^{\scriptsize\blacktriangle}$ & 40.40 & 55.70$^{\scriptsize\blacktriangle}$ \\
    & Nomic-v2 & 0.5B & 17.90 & 41.20$^{\scriptsize\blacktriangle}$ & 35.40 & 65.80$^{\scriptsize\blacktriangle}$ & 24.20 & 45.00$^{\scriptsize\blacktriangle}$ & 46.60 & 62.50$^{\scriptsize\blacktriangle}$ \\
    & BGE-M3 (Dense) & 0.5B & 22.50 & 30.10$^{\scriptsize\blacktriangle}$ & 42.80 & 52.50$^{\scriptsize\blacktriangle}$ & 22.60 & 37.10$^{\scriptsize\blacktriangle}$ & 42.80 & 55.90$^{\scriptsize\blacktriangle}$ \\
    & Multilingual-E5$_{\text{large-instruct}}$ & 0.6B & 25.60 & 35.60$^{\scriptsize\blacktriangle}$ & 45.80 & 58.70$^{\scriptsize\blacktriangle}$ & 30.10 & 40.20$^{\scriptsize\blacktriangle}$ & 52.40 & 63.30$^{\scriptsize\blacktriangle}$ \\
    & Qwen3-Embedding$_{\text{0.6B}}$ & 0.6B & 15.70 & 28.40$^{\scriptsize\blacktriangle}$ & 30.40 & 45.70$^{\scriptsize\blacktriangle}$ & 25.90 & 31.70$^{\scriptsize\blacktriangle}$ & 45.20 & 47.30$^{\scriptsize\blacktriangle}$ \\
    & InstructOR$_{\text{xl}}$ & 1.5B & 23.60 & \textbf{55.95}$^{\scriptsize\blacktriangle}$ & 42.50 & \textbf{74.55}$^{\scriptsize\blacktriangle}$ & 31.80 & \textbf{59.10}$^{\scriptsize\blacktriangle}$ & 58.40 & 62.3$^{\scriptsize\blacktriangle}$ \\
    & Lychee-embed & 1.5B & 12.40 & 11.20$^{\scriptsize\triangledown}$ & 27.00 & 18.40$^{\scriptsize\triangledown}$ & 20.40 & 15.80$^{\scriptsize\triangledown}$ & 41.20 & 26.90$^{\scriptsize\triangledown}$ \\
    & GTE$_{\text{Qwen2-1.5B}}$ & 1.5B & 16.40 & 21.40$^{\scriptsize\blacktriangle}$ & 33.00 & 33.20$^{\scriptsize\blacktriangle}$ & 48.00 & 50.40$^{\scriptsize\blacktriangle}$ & 69.80 & 73.10$^{\scriptsize\blacktriangle}$ \\
    & Qwen3-Embedding$_{\text{4B}}$ & 4B & 20.00 & 27.00$^{\scriptsize\blacktriangle}$ & 36.40 & 43.70$^{\scriptsize\blacktriangle}$ & 26.50 & 39.00$^{\scriptsize\blacktriangle}$ & 44.90 & 59.20$^{\scriptsize\blacktriangle}$ \\
    & Linq-Embed-Mistral & 7B & 34.60 & 37.60$^{\scriptsize\blacktriangle}$ & 55.00 & 58.30$^{\scriptsize\blacktriangle}$ & \textbf{52.80} & 51.50$^{\scriptsize\triangledown}$ & \textbf{75.50} & 72.50$^{\scriptsize\triangledown}$ \\
    & SFR-Embedding-Mistral & 7B & 31.40 & 33.60$^{\scriptsize\blacktriangle}$ & 50.40 & 53.70$^{\scriptsize\blacktriangle}$ & 45.00 & 44.70$^{\scriptsize\triangledown}$ & 66.40 & 65.50$^{\scriptsize\triangledown}$ \\
    & E5$_{\text{Mistral}}$ & 7B & 31.50 & 32.40$^{\scriptsize\blacktriangle}$ & 52.00 & 53.10$^{\scriptsize\blacktriangle}$ & 42.10 & 44.20$^{\scriptsize\blacktriangle}$ & 64.40 & 64.90$^{\scriptsize\blacktriangle}$ \\
    & GritLM$_{\text{7B}}$ & 7B & 13.90 & 32.80$^{\scriptsize\blacktriangle}$ & 32.90 & 63.00$^{\scriptsize\blacktriangle}$ & 49.50 & 57.40$^{\scriptsize\blacktriangle}$ & 72.90 & 77.10$^{\scriptsize\blacktriangle}$ \\
    & GTE$_{\text{Qwen2-7B}}$ & 7B & 2.90 & 6.00$^{\scriptsize\blacktriangle}$ & 6.60 & 11.20$^{\scriptsize\blacktriangle}$ & 36.40 & 40.10$^{\scriptsize\blacktriangle}$ & 61.70 & 64.00$^{\scriptsize\blacktriangle}$ \\
    & Qwen3-Embedding$_{\text{8B}}$ & 8B & 25.10 & 31.60$^{\scriptsize\blacktriangle}$ & 43.70 & 50.50$^{\scriptsize\blacktriangle}$ & 31.80 & 49.50$^{\scriptsize\blacktriangle}$ & 52.10 & 69.50$^{\scriptsize\blacktriangle}$ \\
    & Llama-Embed$_{\text{Nemotron-8B}}$ & 8B & 23.80 & 41.00$^{\scriptsize\blacktriangle}$ & 41.10 & 62.60$^{\scriptsize\blacktriangle}$ & 41.62 & 54.20$^{\scriptsize\blacktriangle}$ & 66.13 & \textbf{78.53}$^{\scriptsize\blacktriangle}$ \\
    & BGE$_{\text{multilingual-gemma2}}$ & 9B & \textbf{37.50} & 12.70$^{\scriptsize\triangledown}$ & 58.90 & 15.30$^{\scriptsize\triangledown}$ & 38.90 & 28.50$^{\scriptsize\triangledown}$ & 61.60 & 44.30$^{\scriptsize\triangledown}$ \\
    \bottomrule
  \end{tabular}
  }
  \caption{Zero-shot results on the IdioLink dataset. Embedding method is indicated by \texttt{sentence} and \texttt{span}. The best result in each configuration is highlighted in bold. 
  The ``$^\blacktriangle$'' symbol indicates an increase and ``$^\triangledown$'' decrease compared to the respective sentence embedding setting.
  }
  \label{tab:zero_shot}
\end{table*}

\paragraph{No instructions in training.} 
We evaluate models that were not post-trained on queries containing instructions.
Such encoder models include Sentence-BERT (SBERT) \citep{reimers-gurevych-2019-sentence}, SimCSE \citep{gao-etal-2021-simcse}, Contriever \citep{izacard2022unsuperviseddenseinformationretrieval} and E5 \citep{wang2024textembeddingsweaklysupervisedcontrastive}.
We also include decoder models: Stella \citep{zhang2025jasperstelladistillationsota} and DRAMA \citep{ma-etal-2025-drama}.

\paragraph{Instructions in training.} Models that are post-trained with instructions given with the queries.
This ranges from minimal, constant instruction to fully flexible instruction-following embedding models.
We assess encoder models: Contriever-based TART \citep{asai-etal-2023-task}, the instruction-tuned multilingual E5 \citep{wang2024multilinguale5textembeddings}, and Nomic-Embed-v2 \citep{nussbaum2025trainingsparsemixtureexperts}.
BGE models \citep{xiao2024cpackpackedresourcesgeneral} and their multilingual version M3 (Dense) \citep{chen-etal-2024-m3} also fall into this category, although they are trained with only one instruction for each broad task (retrieval, clustering, etc.)
We also evaluate decoder models.: InstructOR \cite{su-etal-2023-one}, GritML-7B \citep{muennighoff2025generativerepresentationalinstructiontuning}, E5-Mistral \citep{wang-etal-2024-improving-text}, the GTE models \citep{li2023generaltextembeddingsmultistage}, which are based on Qwen-2 \citep{yang2024qwen2technicalreport}, the SFR-Embedding-Mistral \citep{SFRAIResearch2024}, and the Linq-Embed-Mistral \citep{choi2024linqembedmistraltechnicalreport}, Lychee-embed \citep{zhang2025phased}, which is based on Qwen-2.5 \citep{qwen2025qwen25technicalreport}, the Gemma-based BGE model \citep{chen-etal-2024-m3}, and the Qwen3 Embedding models \citep{zhang2025qwen3embeddingadvancingtext}.

\section{Results and Discussion}
Table~\ref{tab:zero_shot} summarizes the zero-shot results, from which we derive the following main findings.

\paragraph{First, augmenting queries with instructions substantially improves retrieval performance.}
Across all models, both instruction-tuned and non-instruction models, augmenting the query with explicit instructions that focus the model on the target span leads to consistent gains in both R-Precision and nDCG (+11.21 R-Precision, +14.83 nDCG on average).
For example, GTE$_{\text{Qwen2-7B}}$ improves by 33.5 points in R-Precision and 55.10 in nDCG.
These gains highlight the benefit of explicit in-query semantic guidance for task-specific retrieval. 
Only one model, BGE-M3 (Dense), shows no measurable improvement from instruction augmentation, though its performance does not degrade.

\paragraph{Second, span embedding substantially improves retrieval performance.}
For most models, replacing sentence embedding with span embedding yields large improvements in both R-Precision and nDCG (+12.13 and +13.49 on average, respectively).
For example, InstructOR$_{\text{xl}}$ gains 32.35 points in R-Precision and 32.05 in nDCG.
This highlights the importance of modeling the localized span that carries the intended meaning for figurative language retrieval.
Only BGE$_{\text{multilingual-gemma2}}$ and Lychee-embed show notable drops under span embedding, indicating this approach does not benefit all embedding architectures equally.

\paragraph{Third, combining instructions and span embedding yields the strongest gains.}
Adding instructions and using span embedding yields the largest gains (+21.99 R-Precision, +22.71 nDCG on average).
For example, GritLM$_{\text{7B}}$ improves by 43.50 points in R-Precision and achieves substantial gains in nDCG.
Models that perform poorly when using span embeddings in isolation also underperform in a combined setting, though the degradation is notably smaller.

\begin{figure*}[ht]
    \centering
    \includegraphics[width=0.55\textwidth]{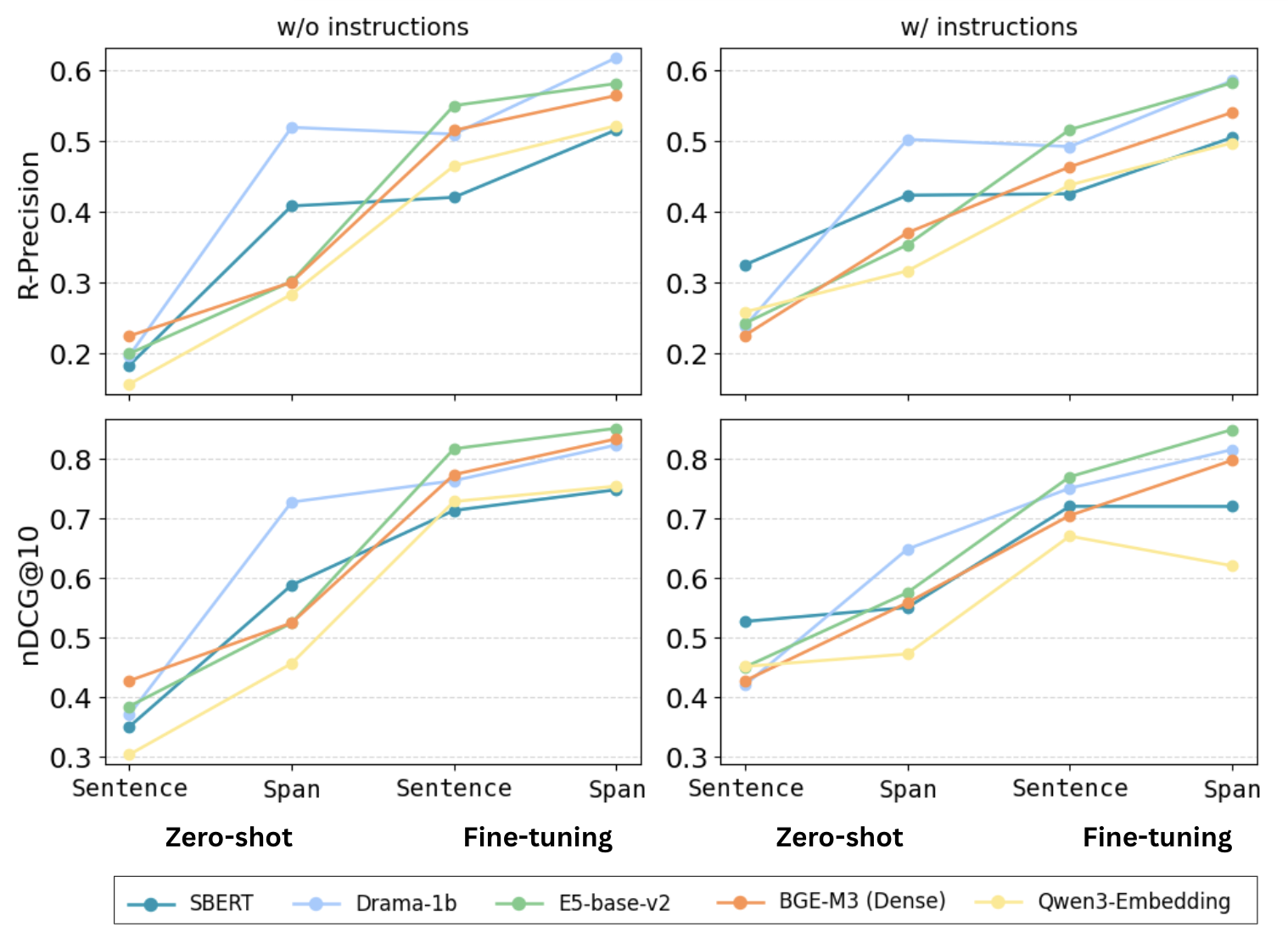}
    \caption{Performance across our four inference-time configurations, where fine-tuning is done with no instructions and with sentence embeddings (\S~\ref{sec:benchmarking}).}
    \label{fig:ft_plot}
\end{figure*}

\paragraph{Fourth, model scale alone does not determine retrieval performance.}
Although larger models often perform well, parameter count alone is not a reliable predictor of retrieval quality.
Smaller or mid-sized models can outperform larger counterparts when paired with effective representation strategies.
For example, across configurations, the 1B Drama$_{\text{1B}}$ achieves R-Precision and nDCG scores of 52.0 and 72.9, respectively, whereas the 9B BGE$_{\text{multilingual-gemma2}}$ reaches only 38.9 and 61.6.
Similarly, the 1.5B InstructOR$_{\text{xl}}$ reaches an R-Precision of 29.1, making it the best-performing model on this metric in the zero-shot setting, and achieves an nDCG of 74.55 in another configuration.
Overall, differences attributable to query formulation and embedding extraction frequently exceed those attributable to model size, highlighting the central role of representation design.


\subsection{Fine-tuning Results}
\label{ssec:ft_res}

Figure~\ref{fig:ft_plot} compares fine-tuned models with their zero-shot counterparts across all four inference-time configurations, while the full results are reported in Table~\ref{tab:ft_full_results} (Appendix~\ref{apdx:ft_apdx}).
Each fine-tuning experiment is repeated using three different random seeds, and results are reported as the mean with the corresponding standard deviation.
Each subplot corresponds to a specific inference-time configuration (\S~\ref{sec:benchmarking}).
For example, when span embedding is used at inference time, fine-tuning is performed either with sentence or with span embedding; in the plot, we report the higher of the two.

\paragraph{Fine-tuning consistently improves performance over zero-shot settings.}
Consistent with the zero-shot findings, span embedding at inference time yields clear performance gains for fine-tuned models, regardless of the training configuration.
In contrast, instructions at inference time typically leads to a slight performance degradation, though it is not substantial.

\paragraph{Training configuration benefits from standard methods.}
Unlike in the zero-shot setting, incorporating instructions during fine-tuning consistently harms performance compared to using raw queries.
Similarly, except for BGE-M3 (Dense), models fine-tuned using span embedding underperform those trained with the standard sentence embedding, assuming span embedding is applied at inference time for a fair comparison.
Overall, the most effective fine-tuning configuration relies on standard methods, queries with no instructions, and sentence embedding.

\section{Conclusions}
We introduced IdioLink, a retrieval benchmark that tests whether embedding models can retrieve conceptually equivalent meanings across literal and idiomatic expressions. 
Our results show that IdioLink is challenging even for state-of-the-art models, with performance driven more by representation choices than by model scale.
In zero-shot settings, both instruction augmentation and span embedding substantially improve retrieval, and their combination yields the strongest results. 
Fine-tuning further improves performance, but only under standard query and embedding configurations; adding instructions or using span embedding in training proved harmful. 
At inference time, however, span embedding provides robust gains, highlighting the importance of decoupling training and inference strategies.

Overall, our findings suggest that effective idiom retrieval requires careful design of representations beyond scaling alone. 
IdioLink offers a controlled testbed for studying conceptual retrieval under semantic variation and opens avenues for future work on meaning-centric evaluation and figurative language understanding.

\section*{Limitations}

IdioLink focuses on English and on a curated set of \acp{PIE}, enabling controlled experimentation but limiting linguistic diversity and cross-lingual generalization. 
Future work could extend the benchmark to additional languages, broader figurative phenomena, and more abstract semantic relations. 
In addition, we collapse multiple figurative senses of the same \ac{PIE} into a single retrieval target, which may introduce ambiguity for some expressions.

Although the dataset undergoes both automatic and human validation, part of the corpus is synthetically generated. 
As with other LLM-generated benchmarks, some artifacts or unnatural distributions may remain despite filtering and manual correction.

Our evaluation is conducted in a controlled retrieval setting with relatively short documents and limited training data (440 training queries). 
Real-world retrieval systems often involve longer documents, noisier inputs, multi-hop reasoning, and larger-scale supervision. 
Similarly, experiments are primarily designed to compare architectures and retrieval configurations rather than optimize individual models exhaustively.

More broadly, recent work suggests that models may exploit dataset-specific shortcuts in synthetic evaluation settings rather than demonstrate robust semantic understanding \citep{hagström2026cubbenchmarkingcontextutilisation}. 
Accordingly, IdioLink should be viewed as a controlled benchmark for abstraction-sensitive semantic retrieval rather than a definitive test of idiom understanding.

Despite these limitations, we believe IdioLink provides a challenging and complementary testbed for evaluating retrieval beyond surface-level similarity.

\section*{Ethics Statement}
\label{Ethics Statement}
We use publicly available datasets and models in accordance with their intended use, as detailed by the respective publishers and under their licenses. 
details are provided in Table~\ref{tab:checkpoints} (Appendix~\ref{apdx:checkpoints}) and in Table~\ref{apdx:artifacts} (Appendix~\ref{apdx:artifacts}).
No personally identifiable information or offensive data are processed, and annotators ensure that none exists in the released dataset.
Our work is intended for research purposes only, and we see no potential risks.

\section*{Acknowledgments}
\label{sec:acknowledgments}
This study is supported in part by the European Research Council (Intellexus, Project No.\@ 101118558).
Views and opinions expressed are, however, those of the authors only and do not necessarily reflect those of the European Union or the European Research Council Executive Agency. Neither the European Union nor the granting authorities can be held responsible for them.

\bibliography{anthology-1,anthology-2,custom}

\begin{thebibliography}{73}
\providecommand{\natexlab}[1]{#1}

\bibitem[{Arslan et~al.(2025)Arslan, {\c{C}}akmak, Eryi{\u{g}}it, and Nivre}]{arslan-etal-2025-using}
Do{\u{g}}ukan Arslan, H{\"u}seyin~An{\i}l {\c{C}}akmak, G{\"u}l{\c{s}}en Eryi{\u{g}}it, and Joakim Nivre. 2025.
\newblock \href {https://doi.org/10.18653/v1/2025.mwe-1.4} {Using {LLM}s to advance idiom corpus construction}.
\newblock In \emph{Proceedings of the 21st Workshop on Multiword Expressions (MWE 2025)}, pages 21--31, Albuquerque, New Mexico, U.S.A. Association for Computational Linguistics.

\bibitem[{Asai et~al.(2023)Asai, Schick, Lewis, Chen, Izacard, Riedel, Hajishirzi, and Yih}]{asai-etal-2023-task}
Akari Asai, Timo Schick, Patrick Lewis, Xilun Chen, Gautier Izacard, Sebastian Riedel, Hannaneh Hajishirzi, and Wen-tau Yih. 2023.
\newblock \href {https://doi.org/10.18653/v1/2023.findings-acl.225} {Task-aware retrieval with instructions}.
\newblock In \emph{Findings of the Association for Computational Linguistics: ACL 2023}, pages 3650--3675, Toronto, Canada. Association for Computational Linguistics.

\bibitem[{Aslanta{\c{s}} and Gungor(2026)}]{aslantas-gungor-2026-unified}
G{\"o}zde Aslanta{\c{s}} and Tunga Gungor. 2026.
\newblock \href {https://doi.org/10.18653/v1/2026.sigturk-1.4} {A unified {T}urkic idiom understanding benchmark: Idiom detection and semantic retrieval across five {T}urkic languages}.
\newblock In \emph{Proceedings of the Second Workshop Natural Language Processing for {T}urkic Languages ({SIGTURK} 2026)}, pages 38--51, Rabat, Morocco. Association for Computational Linguistics.

\bibitem[{Briskilal and Subalalitha(2022)}]{BRISKILAL2022102756}
J~Briskilal and C.N. Subalalitha. 2022.
\newblock \href {https://doi.org/10.1016/j.ipm.2021.102756} {An ensemble model for classifying idioms and literal texts using {BERT} and {RoBERTa}}.
\newblock \emph{Information Processing \& Management}, 59(1):102756.

\bibitem[{Chen et~al.(2024)Chen, Xiao, Zhang, Luo, Lian, and Liu}]{chen-etal-2024-m3}
Jianlyu Chen, Shitao Xiao, Peitian Zhang, Kun Luo, Defu Lian, and Zheng Liu. 2024.
\newblock \href {https://doi.org/10.18653/v1/2024.findings-acl.137} {{M}3-embedding: Multi-linguality, multi-functionality, multi-granularity text embeddings through self-knowledge distillation}.
\newblock In \emph{Findings of the Association for Computational Linguistics: ACL 2024}, pages 2318--2335, Bangkok, Thailand. Association for Computational Linguistics.

\bibitem[{Choi et~al.(2024)Choi, Kim, Lee, Kwon, Gu, Kim, Cho, and yong Sohn}]{choi2024linqembedmistraltechnicalreport}
Chanyeol Choi, Junseong Kim, Seolhwa Lee, Jihoon Kwon, Sangmo Gu, Yejin Kim, Minkyung Cho, and Jy~yong Sohn. 2024.
\newblock \href {https://arxiv.org/abs/2412.03223} {{Linq-Embed-Mistral} technical report}.
\newblock \emph{Preprint}, arXiv:2412.03223.

\bibitem[{Constant et~al.(2017)Constant, Eryiǧit, Monti, van~der Plas, Ramisch, Rosner, and Todirascu}]{constant-etal-2017-survey}
Mathieu Constant, G{\"u}l{\c{s}}en Eryiǧit, Johanna Monti, Lonneke van~der Plas, Carlos Ramisch, Michael Rosner, and Amalia Todirascu. 2017.
\newblock \href {https://doi.org/10.1162/COLI_a_00302} {{S}urvey: Multiword expression processing: A {S}urvey}.
\newblock \emph{Computational Linguistics}, 43(4):837--892.

\bibitem[{Cook et~al.(2007)Cook, Fazly, and Stevenson}]{cook-etal-2007-pulling}
Paul Cook, Afsaneh Fazly, and Suzanne Stevenson. 2007.
\newblock \href {https://aclanthology.org/W07-1106/} {Pulling their weight: Exploiting syntactic forms for the automatic identification of idiomatic expressions in context}.
\newblock In \emph{Proceedings of the Workshop on A Broader Perspective on Multiword Expressions}, pages 41--48, Prague, Czech Republic. Association for Computational Linguistics.

\bibitem[{Cook et~al.(2008)Cook, Fazly, and Stevenson}]{cook2008vnctokens}
Paul Cook, Afsaneh Fazly, and Suzanne Stevenson. 2008.
\newblock The {VNC}-{T}okens dataset.
\newblock In \emph{Proceedings of the LREC Workshop Towards a Shared Task for Multiword Expressions}, pages 19--22.

\bibitem[{De~Luca~Fornaciari et~al.(2024)De~Luca~Fornaciari, Altuna, Gonzalez-Dios, and Melero}]{de-luca-fornaciari-etal-2024-hard}
Francesca De~Luca~Fornaciari, Bego{\~n}a Altuna, Itziar Gonzalez-Dios, and Maite Melero. 2024.
\newblock \href {https://doi.org/10.18653/v1/2024.figlang-1.5} {A hard nut to crack: Idiom detection with conversational large language models}.
\newblock In \emph{Proceedings of the 4th Workshop on Figurative Language Processing (FigLang 2024)}, pages 35--44, Mexico City, Mexico (Hybrid). Association for Computational Linguistics.

\bibitem[{Eryiğit et~al.(2022)Eryiğit, Şentaş, and Monti}]{Eryi_it_2022}
GülŞen Eryiğit, Ali Şentaş, and Johanna Monti. 2022.
\newblock \href {https://doi.org/10.1017/s1351324921000401} {Gamified crowdsourcing for idiom corpora construction}.
\newblock \emph{Natural Language Engineering}, 29(4):909–941.

\bibitem[{Fazly et~al.(2009)Fazly, Cook, and Stevenson}]{fazly-etal-2009-unsupervised}
Afsaneh Fazly, Paul Cook, and Suzanne Stevenson. 2009.
\newblock \href {https://doi.org/10.1162/coli.08-010-R1-07-048} {Unsupervised type and token identification of idiomatic expressions}.
\newblock \emph{Computational Linguistics}, 35(1):61--103.

\bibitem[{Gao et~al.(2021)Gao, Yao, and Chen}]{gao-etal-2021-simcse}
Tianyu Gao, Xingcheng Yao, and Danqi Chen. 2021.
\newblock \href {https://doi.org/10.18653/v1/2021.emnlp-main.552} {{S}im{CSE}: Simple contrastive learning of sentence embeddings}.
\newblock In \emph{Proceedings of the 2021 Conference on Empirical Methods in Natural Language Processing}, pages 6894--6910, Online and Punta Cana, Dominican Republic. Association for Computational Linguistics.

\bibitem[{Gharbieh et~al.(2016)Gharbieh, Bhavsar, and Cook}]{gharbieh-etal-2016-word}
Waseem Gharbieh, Virendra Bhavsar, and Paul Cook. 2016.
\newblock \href {https://doi.org/10.18653/v1/W16-1817} {A word embedding approach to identifying verb-noun idiomatic combinations}.
\newblock In \emph{Proceedings of the 12th Workshop on Multiword Expressions}, pages 112--118, Berlin, Germany. Association for Computational Linguistics.

\bibitem[{Günther et~al.(2025)Günther, Mohr, Williams, Wang, and Xiao}]{günther2025latechunkingcontextualchunk}
Michael Günther, Isabelle Mohr, Daniel~James Williams, Bo~Wang, and Han Xiao. 2025.
\newblock \href {https://arxiv.org/abs/2409.04701} {Late chunking: Contextual chunk embeddings using long-context embedding models}.
\newblock \emph{Preprint}, arXiv:2409.04701.

\bibitem[{Haagsma et~al.(2020)Haagsma, Bos, and Nissim}]{haagsma-etal-2020-magpie}
Hessel Haagsma, Johan Bos, and Malvina Nissim. 2020.
\newblock \href {https://aclanthology.org/2020.lrec-1.35/} {{MAGPIE}: A large corpus of potentially idiomatic expressions}.
\newblock In \emph{Proceedings of the Twelfth Language Resources and Evaluation Conference}, pages 279--287, Marseille, France. European Language Resources Association.

\bibitem[{Hagström et~al.(2026)Hagström, Kim, Yu, goo Lee, Johansson, Cho, and Augenstein}]{hagström2026cubbenchmarkingcontextutilisation}
Lovisa Hagström, Youna Kim, Haeun Yu, Sang goo Lee, Richard Johansson, Hyunsoo Cho, and Isabelle Augenstein. 2026.
\newblock \href {https://arxiv.org/abs/2505.16518} {{CUB}: Benchmarking context utilisation techniques for language models}.
\newblock \emph{Preprint}, arXiv:2505.16518.

\bibitem[{Han et~al.(2020)Han, Jones, and Smeaton}]{han-etal-2020-alphamwe}
Lifeng Han, Gareth Jones, and Alan Smeaton. 2020.
\newblock \href {https://aclanthology.org/2020.mwe-1.6/} {{A}lpha{MWE}: Construction of multilingual parallel corpora with {MWE} annotations}.
\newblock In \emph{Proceedings of the Joint Workshop on Multiword Expressions and Electronic Lexicons}, pages 44--57, online. Association for Computational Linguistics.

\bibitem[{Hasan and Ng(2014)}]{hasan-ng-2014-automatic}
Kazi~Saidul Hasan and Vincent Ng. 2014.
\newblock \href {https://doi.org/10.3115/v1/P14-1119} {Automatic keyphrase extraction: A survey of the state of the art}.
\newblock In \emph{Proceedings of the 52nd Annual Meeting of the Association for Computational Linguistics (Volume 1: Long Papers)}, pages 1262--1273, Baltimore, Maryland. Association for Computational Linguistics.

\bibitem[{Hashiloni et~al.(2025)Hashiloni, Hefetz, and Bar}]{hashiloni-etal-2025-easy}
Kai~Golan Hashiloni, Ofri Hefetz, and Kfir Bar. 2025.
\newblock \href {https://doi.org/10.18653/v1/2025.emnlp-main.1213} {Easy as {PIE}? identifying multi-word expressions with {LLM}s}.
\newblock In \emph{Proceedings of the 2025 Conference on Empirical Methods in Natural Language Processing}, pages 23771--23790, Suzhou, China. Association for Computational Linguistics.

\bibitem[{He et~al.(2024)He, Idiart, Scarton, and Villavicencio}]{he-etal-2024-enhancing}
Wei He, Marco Idiart, Carolina Scarton, and Aline Villavicencio. 2024.
\newblock \href {https://doi.org/10.18653/v1/2024.findings-acl.741} {Enhancing idiomatic representation in multiple languages via an adaptive contrastive triplet loss}.
\newblock In \emph{Findings of the Association for Computational Linguistics: ACL 2024}, pages 12473--12485, Bangkok, Thailand. Association for Computational Linguistics.

\bibitem[{Izacard et~al.(2022)Izacard, Caron, Hosseini, Riedel, Bojanowski, Joulin, and Grave}]{izacard2022unsuperviseddenseinformationretrieval}
Gautier Izacard, Mathilde Caron, Lucas Hosseini, Sebastian Riedel, Piotr Bojanowski, Armand Joulin, and Edouard Grave. 2022.
\newblock \href {https://arxiv.org/abs/2112.09118} {Unsupervised dense information retrieval with contrastive learning}.
\newblock \emph{Preprint}, arXiv:2112.09118.

\bibitem[{Jackendoff(1997)}]{jackendoff1997architecture}
Ray~S. Jackendoff. 1997.
\newblock \emph{The Architecture of the Language Faculty}, volume~28 of \emph{Linguistic Inquiry Monographs}.
\newblock MIT Press, Cambridge, MA; London, England.

\bibitem[{Jha et~al.(2024)Jha, Wang, G{\"u}nther, Mastrapas, Sturua, Mohr, Koukounas, Akram, Wang, and Xiao}]{xiao-etal-2024-jina}
Rohan Jha, Bo~Wang, Michael G{\"u}nther, Georgios Mastrapas, Saba Sturua, Isabelle Mohr, Andreas Koukounas, Mohammad~Kalim Akram, Nan Wang, and Han Xiao. 2024.
\newblock \href {https://doi.org/10.18653/v1/2024.mrl-1.11} {{J}ina-{C}ol{BERT}-v2: A general-purpose multilingual late interaction retriever}.
\newblock In \emph{Proceedings of the Fourth Workshop on Multilingual Representation Learning (MRL 2024)}, pages 159--166, Miami, Florida, USA. Association for Computational Linguistics.

\bibitem[{Kamradt(2024)}]{kamradt2024fivelevels}
Greg Kamradt. 2024.
\newblock \href {https://github.com/FullStackRetrieval-com/RetrievalTutorials/blob/main/tutorials/LevelsOfTextSplitting/5_Levels_Of_Text_Splitting.ipynb} {5 levels of text splitting}.
\newblock GitHub repository.
\newblock Accessed: 2025-12-27.

\bibitem[{Kanerva et~al.(2025)Kanerva, Kitti, Chang, Vahtola, Creutz, and Ginter}]{Kanerva2025}
Jenna Kanerva, Hanna Kitti, Li-Hsin Chang, Teemu Vahtola, Mathias Creutz, and Filip Ginter. 2025.
\newblock \href {https://doi.org/10.1007/s10579-023-09715-7} {Semantic search as extractive paraphrase span detection}.
\newblock \emph{Language Resources and Evaluation}, 59(1):257--276.

\bibitem[{Karpukhin et~al.(2020)Karpukhin, Oguz, Min, Lewis, Wu, Edunov, Chen, and Yih}]{karpukhin-etal-2020-dense}
Vladimir Karpukhin, Barlas Oguz, Sewon Min, Patrick Lewis, Ledell Wu, Sergey Edunov, Danqi Chen, and Wen-tau Yih. 2020.
\newblock \href {https://doi.org/10.18653/v1/2020.emnlp-main.550} {Dense passage retrieval for open-domain question answering}.
\newblock In \emph{Proceedings of the 2020 Conference on Empirical Methods in Natural Language Processing (EMNLP)}, pages 6769--6781, Online. Association for Computational Linguistics.

\bibitem[{Kim et~al.(2025)Kim, Shin, Hwang, Choi, Xuan, and Kim}]{kim-etal-2025-memorization}
Jisu Kim, Youngwoo Shin, Uiji Hwang, Jihun Choi, Richeng Xuan, and Taeuk Kim. 2025.
\newblock \href {https://doi.org/10.18653/v1/2025.emnlp-main.1099} {Memorization or reasoning? exploring the idiom understanding of {LLM}s}.
\newblock In \emph{Proceedings of the 2025 Conference on Empirical Methods in Natural Language Processing}, pages 21678--21699, Suzhou, China. Association for Computational Linguistics.

\bibitem[{Korkontzelos et~al.(2013)Korkontzelos, Zesch, Zanzotto, and Biemann}]{korkontzelos-etal-2013-semeval}
Ioannis Korkontzelos, Torsten Zesch, Fabio~Massimo Zanzotto, and Chris Biemann. 2013.
\newblock \href {https://aclanthology.org/S13-2007/} {{S}em{E}val-2013 task 5: Evaluating phrasal semantics}.
\newblock In \emph{Second Joint Conference on Lexical and Computational Semantics (*{SEM}), Volume 2: Proceedings of the Seventh International Workshop on Semantic Evaluation ({S}em{E}val 2013)}, pages 39--47, Atlanta, Georgia, USA. Association for Computational Linguistics.

\bibitem[{Lee et~al.(2024)Lee, Dai, Ren, Chen, Cer, Cole, Hui, Boratko, Kapadia, Ding, Luan, Duddu, Abrego, Shi, Gupta, Kusupati, Jain, Jonnalagadda, Chang, and Naim}]{lee2024geckoversatiletextembeddings}
Jinhyuk Lee, Zhuyun Dai, Xiaoqi Ren, Blair Chen, Daniel Cer, Jeremy~R. Cole, Kai Hui, Michael Boratko, Rajvi Kapadia, Wen Ding, Yi~Luan, Sai Meher~Karthik Duddu, Gustavo~Hernandez Abrego, Weiqiang Shi, Nithi Gupta, Aditya Kusupati, Prateek Jain, Siddhartha~Reddy Jonnalagadda, Ming-Wei Chang, and Iftekhar Naim. 2024.
\newblock \href {https://arxiv.org/abs/2403.20327} {Gecko: Versatile text embeddings distilled from large language models}.
\newblock \emph{Preprint}, arXiv:2403.20327.

\bibitem[{Lewis et~al.(2021)Lewis, Perez, Piktus, Petroni, Karpukhin, Goyal, Küttler, Lewis, tau Yih, Rocktäschel, Riedel, and Kiela}]{lewis2021retrievalaugmentedgenerationknowledgeintensivenlp}
Patrick Lewis, Ethan Perez, Aleksandra Piktus, Fabio Petroni, Vladimir Karpukhin, Naman Goyal, Heinrich Küttler, Mike Lewis, Wen tau Yih, Tim Rocktäschel, Sebastian Riedel, and Douwe Kiela. 2021.
\newblock \href {https://arxiv.org/abs/2005.11401} {Retrieval-augmented generation for knowledge-intensive {NLP} tasks}.
\newblock \emph{Preprint}, arXiv:2005.11401.

\bibitem[{Li et~al.(2023)Li, Zhang, Zhang, Long, Xie, and Zhang}]{li2023generaltextembeddingsmultistage}
Zehan Li, Xin Zhang, Yanzhao Zhang, Dingkun Long, Pengjun Xie, and Meishan Zhang. 2023.
\newblock \href {https://arxiv.org/abs/2308.03281} {Towards general text embeddings with multi-stage contrastive learning}.
\newblock \emph{Preprint}, arXiv:2308.03281.

\bibitem[{Ma et~al.(2025)Ma, Lin, Oguz, Lin, Yih, and Chen}]{ma-etal-2025-drama}
Xueguang Ma, Xi~Victoria Lin, Barlas Oguz, Jimmy Lin, Wen-tau Yih, and Xilun Chen. 2025.
\newblock \href {https://doi.org/10.18653/v1/2025.acl-long.1457} {{DRAMA}: Diverse augmentation from large language models to smaller dense retrievers}.
\newblock In \emph{Proceedings of the 63rd Annual Meeting of the Association for Computational Linguistics (Volume 1: Long Papers)}, pages 30170--30186, Vienna, Austria. Association for Computational Linguistics.

\bibitem[{Meng et~al.(2024)Meng, Liu, Joty, Xiong, Zhou, and Yavuz}]{SFRAIResearch2024}
Rui Meng, Ye~Liu, Shafiq~Rayhan Joty, Caiming Xiong, Yingbo Zhou, and Semih Yavuz. 2024.
\newblock \href {https://www.salesforce.com/blog/sfr-embedding/} {{SFR-Embedding-Mistral}:enhance text retrieval with transfer learning}.
\newblock Salesforce AI Research Blog.

\bibitem[{Mi et~al.(2025)Mi, Villavicencio, and Moosavi}]{mi-etal-2025-rolling}
Maggie Mi, Aline Villavicencio, and Nafise~Sadat Moosavi. 2025.
\newblock \href {https://doi.org/10.18653/v1/2025.acl-long.362} {Rolling the {DICE} on idiomaticity: How {LLM}s fail to grasp context}.
\newblock In \emph{Proceedings of the 63rd Annual Meeting of the Association for Computational Linguistics (Volume 1: Long Papers)}, pages 7314--7332, Vienna, Austria. Association for Computational Linguistics.

\bibitem[{Miao et~al.(2023)Miao, Du, and Zhang}]{Miao_2023}
Pu~Miao, Zeyao Du, and Junlin Zhang. 2023.
\newblock \href {https://doi.org/10.1145/3583780.3614833} {Deb{CSE}: Rethinking unsupervised contrastive sentence embedding learning in the debiasing perspective}.
\newblock In \emph{Proceedings of the 32nd ACM International Conference on Information and Knowledge Management}, (CIKM 23), page 1847–1856. ACM.

\bibitem[{Muennighoff et~al.(2025)Muennighoff, Su, Wang, Yang, Wei, Yu, Singh, and Kiela}]{muennighoff2025generativerepresentationalinstructiontuning}
Niklas Muennighoff, Hongjin Su, Liang Wang, Nan Yang, Furu Wei, Tao Yu, Amanpreet Singh, and Douwe Kiela. 2025.
\newblock \href {https://arxiv.org/abs/2402.09906} {Generative representational instruction tuning}.
\newblock \emph{Preprint}, arXiv:2402.09906.

\bibitem[{Nussbaum and Duderstadt(2025)}]{nussbaum2025trainingsparsemixtureexperts}
Zach Nussbaum and Brandon Duderstadt. 2025.
\newblock \href {https://arxiv.org/abs/2502.07972} {Training sparse mixture of experts text embedding models}.
\newblock \emph{Preprint}, arXiv:2502.07972.

\bibitem[{Park et~al.(2025)Park, Choi, Kim, An, Wang, Choi, and Kim}]{park-etal-2025-fluid}
Seoyoon Park, Hyeji Choi, Minseon Kim, Subin An, Xiaonan Wang, Gyuri Choi, and Hansaem Kim. 2025.
\newblock \href {https://doi.org/10.18653/v1/2025.emnlp-main.1540} {{FLUID} {QA}: A multilingual benchmark for figurative language usage in dialogue across {E}nglish, {C}hinese, and {K}orean}.
\newblock In \emph{Proceedings of the 2025 Conference on Empirical Methods in Natural Language Processing}, pages 30280--30294, Suzhou, China. Association for Computational Linguistics.

\bibitem[{Phelps et~al.(2024)Phelps, Pickard, Mi, Gow-Smith, and Villavicencio}]{phelps-etal-2024-sign}
Dylan Phelps, Thomas Pickard, Maggie Mi, Edward Gow-Smith, and Aline Villavicencio. 2024.
\newblock \href {https://aclanthology.org/2024.mwe-1.22/} {Sign of the times: Evaluating the use of large language models for idiomaticity detection}.
\newblock In \emph{Proceedings of the Joint Workshop on Multiword Expressions and Universal Dependencies (MWE-UD) @ LREC-COLING 2024}, pages 178--187, Torino, Italia. ELRA and ICCL.

\bibitem[{Pollio et~al.(1977)Pollio, Barlow, Fine, and Pollio}]{pollio1977psychology}
Howard~R. Pollio, John~M. Barlow, Howard~J. Fine, and Marilyn~R. Pollio. 1977.
\newblock \emph{Psychology and the Poetics of Growth: Figurative Language in Psychology, Psychotherapy, and Education}.
\newblock Lawrence Erlbaum, Hillsdale, NJ.

\bibitem[{Qiang et~al.(2023)Qiang, Li, Zhang, Li, Zhu, Yuan, and Wu}]{qiang-etal-2023-chinese-idiom}
Jipeng Qiang, Yang Li, Chaowei Zhang, Yun Li, Yi~Zhu, Yunhao Yuan, and Xindong Wu. 2023.
\newblock \href {https://doi.org/10.1162/tacl_a_00572} {{C}hinese idiom paraphrasing}.
\newblock \emph{Transactions of the Association for Computational Linguistics}, 11:740--754.

\bibitem[{Qwen et~al.(2025)Qwen, :, Yang, Yang, Zhang, Hui, Zheng, Yu, Li, Liu, Huang, Wei, Lin, Yang, Tu, Zhang, Yang, Yang, Zhou, Lin, Dang, Lu, Bao, Yang, Yu, Li, Xue, Zhang, Zhu, Men, Lin, Li, Tang, Xia, Ren, Ren, Fan, Su, Zhang, Wan, Liu, Cui, Zhang, and Qiu}]{qwen2025qwen25technicalreport}
Qwen, :, An~Yang, Baosong Yang, Beichen Zhang, Binyuan Hui, Bo~Zheng, Bowen Yu, Chengyuan Li, Dayiheng Liu, Fei Huang, Haoran Wei, Huan Lin, Jian Yang, Jianhong Tu, Jianwei Zhang, Jianxin Yang, Jiaxi Yang, Jingren Zhou, and 25 others. 2025.
\newblock \href {https://arxiv.org/abs/2412.15115} {Qwen2.5 technical report}.
\newblock \emph{Preprint}, arXiv:2412.15115.

\bibitem[{Ramisch et~al.(2020)Ramisch, Savary, Guillaume, Waszczuk, Candito, Vaidya, Barbu~Mititelu, Bhatia, I{\~n}urrieta, Giouli, G{\"u}ng{\"o}r, Jiang, Lichte, Liebeskind, Monti, Ramisch, Stymne, Walsh, and Xu}]{ramisch-etal-2020-edition}
Carlos Ramisch, Agata Savary, Bruno Guillaume, Jakub Waszczuk, Marie Candito, Ashwini Vaidya, Verginica Barbu~Mititelu, Archna Bhatia, Uxoa I{\~n}urrieta, Voula Giouli, Tunga G{\"u}ng{\"o}r, Menghan Jiang, Timm Lichte, Chaya Liebeskind, Johanna Monti, Renata Ramisch, Sara Stymne, Abigail Walsh, and Hongzhi Xu. 2020.
\newblock \href {https://aclanthology.org/2020.mwe-1.14/} {Edition 1.2 of the {PARSEME} shared task on semi-supervised identification of verbal multiword expressions}.
\newblock In \emph{Proceedings of the Joint Workshop on Multiword Expressions and Electronic Lexicons}, pages 107--118, online. Association for Computational Linguistics.

\bibitem[{Reimers and Gurevych(2019)}]{reimers-gurevych-2019-sentence}
Nils Reimers and Iryna Gurevych. 2019.
\newblock \href {https://doi.org/10.18653/v1/D19-1410} {Sentence-{BERT}: Sentence embeddings using {S}iamese {BERT}-networks}.
\newblock In \emph{Proceedings of the 2019 Conference on Empirical Methods in Natural Language Processing and the 9th International Joint Conference on Natural Language Processing (EMNLP-IJCNLP)}, pages 3982--3992, Hong Kong, China. Association for Computational Linguistics.

\bibitem[{Robertson and Zaragoza(2009)}]{bm25}
Stephen Robertson and Hugo Zaragoza. 2009.
\newblock \href {https://doi.org/10.1561/1500000019} {The probabilistic relevance framework: Bm25 and beyond}.
\newblock \emph{Found. Trends Inf. Retr.}, 3(4):333–389.

\bibitem[{Santhanam et~al.(2022)Santhanam, Khattab, Saad-Falcon, Potts, and Zaharia}]{santhanam-etal-2022-colbertv2}
Keshav Santhanam, Omar Khattab, Jon Saad-Falcon, Christopher Potts, and Matei Zaharia. 2022.
\newblock \href {https://doi.org/10.18653/v1/2022.naacl-main.272} {{C}ol{BERT}v2: Effective and efficient retrieval via lightweight late interaction}.
\newblock In \emph{Proceedings of the 2022 Conference of the North American Chapter of the Association for Computational Linguistics: Human Language Technologies}, pages 3715--3734, Seattle, United States. Association for Computational Linguistics.

\bibitem[{Savary et~al.(2023)Savary, Ben~Khelil, Ramisch, Giouli, Barbu~Mititelu, Hadj~Mohamed, Krstev, Liebeskind, Xu, Stymne, G{\"u}ng{\"o}r, Pickard, Guillaume, Bej{\v{c}}ek, Bhatia, Candito, Gantar, I{\~n}urrieta, Gatt, Kovalevskaite, Lichte, Ljube{\v{s}}i{\'c}, Monti, Parra~Escart{\'i}n, Shamsfard, Stoyanova, Vincze, and Walsh}]{savary-etal-2023-parseme}
Agata Savary, Cherifa Ben~Khelil, Carlos Ramisch, Voula Giouli, Verginica Barbu~Mititelu, Najet Hadj~Mohamed, Cvetana Krstev, Chaya Liebeskind, Hongzhi Xu, Sara Stymne, Tunga G{\"u}ng{\"o}r, Thomas Pickard, Bruno Guillaume, Eduard Bej{\v{c}}ek, Archna Bhatia, Marie Candito, Polona Gantar, Uxoa I{\~n}urrieta, Albert Gatt, and 9 others. 2023.
\newblock \href {https://doi.org/10.18653/v1/2023.mwe-1.6} {{PARSEME} corpus release 1.3}.
\newblock In \emph{Proceedings of the 19th Workshop on Multiword Expressions (MWE 2023)}, pages 24--35, Dubrovnik, Croatia. Association for Computational Linguistics.

\bibitem[{Saxena and Paul(2020)}]{saxena2020epiedatasetcorpuspossible}
Prateek Saxena and Soma Paul. 2020.
\newblock \href {https://arxiv.org/abs/2006.09479} {{EPIE} dataset: A corpus for possible idiomatic expressions}.
\newblock \emph{Preprint}, arXiv:2006.09479.

\bibitem[{Scholivet et~al.(2026)Scholivet, Savary, Ramisch, Bilinski, Nakamura, Mitrofan, and Pais}]{scholivet-etal-2026-edition}
Manon Scholivet, Agata Savary, Carlos Ramisch, Eric Bilinski, Takuya Nakamura, Maria Mitrofan, and Vasile Pais. 2026.
\newblock \href {https://doi.org/10.18653/v1/2026.mwe-1.33} {Edition 2.0 of the {PARSEME} shared task on multilingual identification and paraphrasing of multiword expressions}.
\newblock In \emph{Proceedings of the 22nd Workshop on Multiword Expressions ({MWE} 2026)}, pages 254--275, Rabat, Marocco. Association for Computational Linguistics.

\bibitem[{Shi et~al.(2023)Shi, Wang, Bai, Li, Li, Cui, Zeng, Chilimbi, and Zhu}]{shi-etal-2023-osscse}
Zhan Shi, Guoyin Wang, Ke~Bai, Jiwei Li, Xiang Li, Qingjun Cui, Belinda Zeng, Trishul Chilimbi, and Xiaodan Zhu. 2023.
\newblock \href {https://doi.org/10.18653/v1/2023.emnlp-main.448} {{O}ss{CSE}: Overcoming surface structure bias in contrastive learning for unsupervised sentence embedding}.
\newblock In \emph{Proceedings of the 2023 Conference on Empirical Methods in Natural Language Processing}, pages 7242--7254, Singapore. Association for Computational Linguistics.

\bibitem[{Sporleder et~al.(2010)Sporleder, Li, Gorinski, and Koch}]{sporleder-etal-2010-idioms}
Caroline Sporleder, Linlin Li, Philip Gorinski, and Xaver Koch. 2010.
\newblock \href {https://aclanthology.org/L10-1425/} {Idioms in context: The {IDIX} corpus}.
\newblock In \emph{Proceedings of the Seventh International Conference on Language Resources and Evaluation ({LREC}'10)}, Valletta, Malta. European Language Resources Association (ELRA).

\bibitem[{Su et~al.(2023)Su, Shi, Kasai, Wang, Hu, Ostendorf, Yih, Smith, Zettlemoyer, and Yu}]{su-etal-2023-one}
Hongjin Su, Weijia Shi, Jungo Kasai, Yizhong Wang, Yushi Hu, Mari Ostendorf, Wen-tau Yih, Noah~A. Smith, Luke Zettlemoyer, and Tao Yu. 2023.
\newblock \href {https://doi.org/10.18653/v1/2023.findings-acl.71} {One embedder, any task: Instruction-finetuned text embeddings}.
\newblock In \emph{Findings of the Association for Computational Linguistics: ACL 2023}, pages 1102--1121, Toronto, Canada. Association for Computational Linguistics.

\bibitem[{Taslimipoor et~al.(2020)Taslimipoor, Bahaadini, and Kochmar}]{taslimipoor-etal-2020-mtlb}
Shiva Taslimipoor, Sara Bahaadini, and Ekaterina Kochmar. 2020.
\newblock \href {https://aclanthology.org/2020.mwe-1.19/} {{MTLB}-{STRUCT} @parseme 2020: Capturing unseen multiword expressions using multi-task learning and pre-trained masked language models}.
\newblock In \emph{Proceedings of the Joint Workshop on Multiword Expressions and Electronic Lexicons}, pages 142--148, online. Association for Computational Linguistics.

\bibitem[{Tayyar~Madabushi et~al.(2022)Tayyar~Madabushi, Gow-Smith, Garcia, Scarton, Idiart, and Villavicencio}]{tayyar-madabushi-etal-2022-semeval}
Harish Tayyar~Madabushi, Edward Gow-Smith, Marcos Garcia, Carolina Scarton, Marco Idiart, and Aline Villavicencio. 2022.
\newblock \href {https://doi.org/10.18653/v1/2022.semeval-1.13} {{S}em{E}val-2022 task 2: Multilingual idiomaticity detection and sentence embedding}.
\newblock In \emph{Proceedings of the 16th International Workshop on Semantic Evaluation (SemEval-2022)}, pages 107--121, Seattle, United States. Association for Computational Linguistics.

\bibitem[{Tedeschi et~al.(2022)Tedeschi, Martelli, and Navigli}]{tedeschi-etal-2022-id10m}
Simone Tedeschi, Federico Martelli, and Roberto Navigli. 2022.
\newblock \href {https://doi.org/10.18653/v1/2022.findings-naacl.208} {{ID}10{M}: Idiom identification in 10 languages}.
\newblock In \emph{Findings of the Association for Computational Linguistics: NAACL 2022}, pages 2715--2726, Seattle, United States. Association for Computational Linguistics.

\bibitem[{Timothy~Baldwin(2010)}]{mwe_nlp_baldwin}
Su~Nam~Kim Timothy~Baldwin. 2010.
\newblock \emph{Handbook of Natural Language Processing}, chapter 2:267-292.

\bibitem[{Wang et~al.(2022)Wang, Li, Huang, Dou, Kong, and Shao}]{wang2022sncsecontrastivelearningunsupervised}
Hao Wang, Yangguang Li, Zhen Huang, Yong Dou, Lingpeng Kong, and Jing Shao. 2022.
\newblock \href {https://arxiv.org/abs/2201.05979} {Sncse: Contrastive learning for unsupervised sentence embedding with soft negative samples}.
\newblock \emph{Preprint}, arXiv:2201.05979.

\bibitem[{Wang et~al.(2024{\natexlab{a}})Wang, Yang, Huang, Jiao, Yang, Jiang, Majumder, and Wei}]{wang2024textembeddingsweaklysupervisedcontrastive}
Liang Wang, Nan Yang, Xiaolong Huang, Binxing Jiao, Linjun Yang, Daxin Jiang, Rangan Majumder, and Furu Wei. 2024{\natexlab{a}}.
\newblock \href {https://arxiv.org/abs/2212.03533} {Text embeddings by weakly-supervised contrastive pre-training}.
\newblock \emph{Preprint}, arXiv:2212.03533.

\bibitem[{Wang et~al.(2024{\natexlab{b}})Wang, Yang, Huang, Yang, Majumder, and Wei}]{wang-etal-2024-improving-text}
Liang Wang, Nan Yang, Xiaolong Huang, Linjun Yang, Rangan Majumder, and Furu Wei. 2024{\natexlab{b}}.
\newblock \href {https://doi.org/10.18653/v1/2024.acl-long.642} {Improving text embeddings with large language models}.
\newblock In \emph{Proceedings of the 62nd Annual Meeting of the Association for Computational Linguistics (Volume 1: Long Papers)}, pages 11897--11916, Bangkok, Thailand. Association for Computational Linguistics.

\bibitem[{Wang et~al.(2024{\natexlab{c}})Wang, Yang, Huang, Yang, Majumder, and Wei}]{wang2024multilinguale5textembeddings}
Liang Wang, Nan Yang, Xiaolong Huang, Linjun Yang, Rangan Majumder, and Furu Wei. 2024{\natexlab{c}}.
\newblock \href {https://arxiv.org/abs/2402.05672} {Multilingual e5 text embeddings: A technical report}.
\newblock \emph{Preprint}, arXiv:2402.05672.

\bibitem[{Wang et~al.(2023)Wang, Wei, Schuurmans, Le, Chi, Narang, Chowdhery, and Zhou}]{wang2023selfconsistencyimproveschainthought}
Xuezhi Wang, Jason Wei, Dale Schuurmans, Quoc Le, Ed~Chi, Sharan Narang, Aakanksha Chowdhery, and Denny Zhou. 2023.
\newblock \href {https://arxiv.org/abs/2203.11171} {Self-{C}onsistency improves chain of thought reasoning in language models}.
\newblock \emph{Preprint}, arXiv:2203.11171.

\bibitem[{Weinreich(1969)}]{weinreich1969problems}
Uriel Weinreich. 1969.
\newblock Problems in the analysis of idioms.
\newblock In \emph{Problems in the Analysis of Idioms}, pages 23--82. University of California Press, Berkeley.

\bibitem[{Weller et~al.(2025)Weller, Chang, MacAvaney, Lo, Cohan, Van~Durme, Lawrie, and Soldaini}]{weller-etal-2025-followir}
Orion Weller, Benjamin Chang, Sean MacAvaney, Kyle Lo, Arman Cohan, Benjamin Van~Durme, Dawn Lawrie, and Luca Soldaini. 2025.
\newblock \href {https://doi.org/10.18653/v1/2025.naacl-long.597} {{F}ollow{IR}: Evaluating and teaching information retrieval models to follow instructions}.
\newblock In \emph{Proceedings of the 2025 Conference of the Nations of the Americas Chapter of the Association for Computational Linguistics: Human Language Technologies (Volume 1: Long Papers)}, pages 11926--11942, Albuquerque, New Mexico. Association for Computational Linguistics.

\bibitem[{Wu et~al.(2024)Wu, Yin, and Chang}]{wu-etal-2024-kpeval}
Di~Wu, Da~Yin, and Kai-Wei Chang. 2024.
\newblock \href {https://doi.org/10.18653/v1/2024.findings-acl.117} {{KPE}val: Towards fine-grained semantic-based keyphrase evaluation}.
\newblock In \emph{Findings of the Association for Computational Linguistics: ACL 2024}, pages 1959--1981, Bangkok, Thailand. Association for Computational Linguistics.

\bibitem[{Xiao et~al.(2024)Xiao, Liu, Zhang, Muennighoff, Lian, and Nie}]{xiao2024cpackpackedresourcesgeneral}
Shitao Xiao, Zheng Liu, Peitian Zhang, Niklas Muennighoff, Defu Lian, and Jian-Yun Nie. 2024.
\newblock \href {https://arxiv.org/abs/2309.07597} {C-pack: Packed resources for general chinese embeddings}.
\newblock \emph{Preprint}, arXiv:2309.07597.

\bibitem[{Yang et~al.(2024)Yang, Yang, Hui, Zheng, Yu, Zhou, Li, Li, Liu, Huang, Dong, Wei, Lin, Tang, Wang, Yang, Tu, Zhang, Ma, Yang, Xu, Zhou, Bai, He, Lin, Dang, Lu, Chen, Yang, Li, Xue, Ni, Zhang, Wang, Peng, Men, Gao, Lin, Wang, Bai, Tan, Zhu, Li, Liu, Ge, Deng, Zhou, Ren, Zhang, Wei, Ren, Liu, Fan, Yao, Zhang, Wan, Chu, Liu, Cui, Zhang, Guo, and Fan}]{yang2024qwen2technicalreport}
An~Yang, Baosong Yang, Binyuan Hui, Bo~Zheng, Bowen Yu, Chang Zhou, Chengpeng Li, Chengyuan Li, Dayiheng Liu, Fei Huang, Guanting Dong, Haoran Wei, Huan Lin, Jialong Tang, Jialin Wang, Jian Yang, Jianhong Tu, Jianwei Zhang, Jianxin Ma, and 43 others. 2024.
\newblock \href {https://arxiv.org/abs/2407.10671} {Qwen2 technical report}.
\newblock \emph{Preprint}, arXiv:2407.10671.

\bibitem[{Zeng and Bhat(2021)}]{zeng-bhat-2021-idiomatic}
Ziheng Zeng and Suma Bhat. 2021.
\newblock \href {https://doi.org/10.1162/tacl_a_00442} {Idiomatic expression identification using semantic compatibility}.
\newblock \emph{Transactions of the Association for Computational Linguistics}, 9:1546--1562.

\bibitem[{Zhang et~al.(2025{\natexlab{a}})Zhang, Li, Zeng, and Wang}]{zhang2025jasperstelladistillationsota}
Dun Zhang, Jiacheng Li, Ziyang Zeng, and Fulong Wang. 2025{\natexlab{a}}.
\newblock \href {https://arxiv.org/abs/2412.19048} {Jasper and stella: distillation of sota embedding models}.
\newblock \emph{Preprint}, arXiv:2412.19048.

\bibitem[{Zhang et~al.(2025{\natexlab{b}})Zhang, Zhang, Xie, Long, Li, Xie, Zhang, Li, and Zhang}]{zhang2025phased}
Xin Zhang, Yanzhao Zhang, Wen Xie, Dingkun Long, Mingxin Li, Pengjun Xie, Meishan Zhang, Wenjie Li, and Min Zhang. 2025{\natexlab{b}}.
\newblock \href {https://openreview.net/forum?id=NC6G1KCxlt} {Phased training for {LLM}-powered text retrieval models beyond data scaling}.
\newblock In \emph{Second Conference on Language Modeling}.

\bibitem[{Zhang et~al.(2025{\natexlab{c}})Zhang, Li, Long, Zhang, Lin, Yang, Xie, Yang, Liu, Lin, Huang, and Zhou}]{zhang2025qwen3embeddingadvancingtext}
Yanzhao Zhang, Mingxin Li, Dingkun Long, Xin Zhang, Huan Lin, Baosong Yang, Pengjun Xie, An~Yang, Dayiheng Liu, Junyang Lin, Fei Huang, and Jingren Zhou. 2025{\natexlab{c}}.
\newblock \href {https://arxiv.org/abs/2506.05176} {Qwen3 embedding: Advancing text embedding and reranking through foundation models}.
\newblock \emph{Preprint}, arXiv:2506.05176.

\bibitem[{Zhou et~al.(2021{\natexlab{a}})Zhou, Gong, and Bhat}]{zhou-etal-2021-pie}
Jianing Zhou, Hongyu Gong, and Suma Bhat. 2021{\natexlab{a}}.
\newblock \href {https://doi.org/10.18653/v1/2021.mwe-1.5} {{PIE}: A parallel idiomatic expression corpus for idiomatic sentence generation and paraphrasing}.
\newblock In \emph{Proceedings of the 17th Workshop on Multiword Expressions (MWE 2021)}, pages 33--48, Online. Association for Computational Linguistics.

\bibitem[{Zhou et~al.(2021{\natexlab{b}})Zhou, Zeng, Gong, and Bhat}]{zhou2021idiomaticexpressionparaphrasingstrong}
Jianing Zhou, Ziheng Zeng, Hongyu Gong, and Suma Bhat. 2021{\natexlab{b}}.
\newblock \href {https://arxiv.org/abs/2112.08592} {Idiomatic expression paraphrasing without strong supervision}.
\newblock \emph{Preprint}, arXiv:2112.08592.

\end{thebibliography}

\appendix

\newpage

\section{Annotators Information}
\label{apdx:annotators_info}

Data annotation and validation are performed by three annotators with academic backgrounds in scientific research: one PhD researcher and two PhD candidates, all of whom are proficient in English.
Annotators volunteered to support the research activities and were thus not paid specifically for performing the annotation task.

\section{Instruction Implementation Details}
\label{apdx:inst}

The instruction is prepended to the original query and formulated as:

\begin{quote}
\small
\texttt{Based on the literal/idiomatic usage of the span \{span\} in the query, retrieve documents that contain a span conveying the same conceptual meaning.}
\end{quote}

where \texttt{\{span\}} denotes the target span in the query.

Instructions are provided in accordance with the recommended usage patterns for each model, as specified by the model's authors or providers.
When available, we follow the model-specific instruction formatting guidelines.
For full implementation details per model, we refer the reader to the project repository.

In practice, we observe two common instruction integration strategies:
\begin{enumerate}
    \item A widely used format adopted by many embedding models, in which the final input is constructed as:
    \begin{quote}
    \small
    \texttt{Instruct: \{instruction\}\\
    Query: \{query\}}
    \end{quote}
   \item Another set of models requires the instruction to be passed separately as an additional argument to the encode function, alongside the queries.
    \begin{quote}
    \small
    \texttt{model.encode(queries, instruction = instruction).}
    \end{quote}

    \item For models that do not support a dedicated instruction field or predefined format, we apply the same instruction-query concatenation method described above.
\end{enumerate}

\section{Technical Details}
\label{apdx:tech_details}
All of our experiments run on an M4 Pro GPU (20C) with 64GB of memory.
Including experimentation and hyperparameter tuning, we estimate that all zero-shot experiments took approximately 24 hours, whereas those with fine-tuning took approximately 90 hours.
For prompting and accessing \acp{LLM}, we use Agno, as it provides a modular, reproducible framework for LLM evaluation, particularly when working with multiple providers, as it wraps their APIs in a unified layer.
We use \textit{SentenceTransformer} to load and use embedding models.
For the fine-tuning approach, we utilize \textit{PyTorch}.

\section{Evaluation Metrics}
\label{apdx:metrics}

We evaluate retrieval quality using R-precision and nDCG@10, two standard information retrieval metrics that measure different aspects of ranking quality.

\paragraph{R-precision.}
For a query $q$ with $K$ relevant documents, R-precision measures the proportion of relevant documents retrieved within the top-$K$ ranked results. 
A perfect score of $1.0$ means that all relevant documents appear within the first $K$ retrieved items.

In our benchmark, the number of relevant documents is 40 for literal questions and 60 for the idiomatic ones

\textbf{Example.}
Suppose a query has $K=5$ relevant documents. 
A model ranks relevant documents at positions $[1, 2, 4, 7, 8]$.
Only the top-5 retrieved documents are considered for R-precision.
Since 3 of the 5 relevant documents appear within the top-5 positions, the R-precision score is:

\[
R\text{-precision} = \frac{3}{5} = 0.6
\]

\paragraph{nDCG@10.}
We also report normalized Discounted Cumulative Gain at rank 10 (nDCG@10), using binary relevance labels. 
Unlike R-precision, nDCG@10 is sensitive to the ordering of retrieved results, assigning greater weight to relevant documents that appear earlier in the ranking. 
This is particularly important in our setting because retrieval models may return partially related documents (e.g., documents that share topical context or lexical overlap) while failing to rank true semantic matches in top positions.

\textbf{Example.}
Consider two models: one ranks these documents at positions $[1,2,3]$ while another ranks them at $[3,7,9]$. 
Both systems retrieve relevant documents, but the first receives a higher nDCG@10 score because semantically correct matches appear earlier in the ranking.

\section{BM25 Baseline}
\label{apdx:bm25}

\subsection{BM25 Setup}
\label{apdx:bm25_setup}

Although IdioLink is designed to evaluate embedding-based semantic retrieval for figurative language, we include BM25 \citep{bm25} as a strong lexical control baseline.
This allows us to estimate how much retrieval performance can be achieved through surface-form overlap alone, without learned semantic representations.

BM25 is a bag-of-words ranking function based on exact term overlap, term frequency saturation, inverse document frequency, and document length normalization.
We use the \texttt{rank-bm25} implementation of BM25Okapi with lowercase regex tokenization (\texttt{\textbackslash b\textbackslash w+(?:'\textbackslash w+)?\textbackslash b}).
All configurations index the \texttt{sentence} field of documents and retrieve the top 100 candidates.

\paragraph{Query configurations.}
We evaluate two BM25 configurations:
\texttt{sentence}, which uses the full query sentence as the retrieval query, and
\texttt{span}, which uses only the annotated PIE span.
The \texttt{span} configuration serves as a task-aware lexical variant that explicitly emphasizes the target expression.

\paragraph{Hyperparameter tuning.}
We tune $k_1$ and $b$ on the validation split using nDCG@10, searching over
$k_1 \in \{0.6, 0.9, 1.2, 1.5, 2.0\}$ and
$b \in \{0.2, 0.4, 0.6, 0.75, 0.9\}$.
Validation selected $k_1=0.9$ and $b=0.4$, matching standard Lucene/Anserini defaults.

\subsection{BM25 Results Analysis}
\label{apdx:bm25_res}

BM25 achieves relatively strong performance due to the controlled structure of IdioLink and the lexical nature of the benchmark.
In the standard \texttt{sentence} configuration, BM25 achieves 59.07 nDCG@10 despite relying solely on lexical overlap, indicating that exact surface-form matching is highly effective for retrieving top-ranked candidates.

This behavior follows directly from the dataset design.
For \texttt{idiomatic} queries, the relevant set contains 60 documents:
20 \texttt{idiomatic}, 20 \texttt{simplification}, and 20 \texttt{sense} documents.
The 20 \texttt{idiomatic} documents contain the exact same PIE as the query, making them highly retrievable by BM25.
At the same time, the 40 \texttt{literal} documents also contain the same PIE and therefore act as lexical distractors despite expressing a different meaning.
Since the \texttt{simplification} and \texttt{sense} documents omit the PIE entirely, BM25 can reliably retrieve only a subset of the relevant set through direct lexical overlap.

A similar effect appears for \texttt{literal} queries, where all 40 relevant \texttt{literal} documents contain the PIE verbatim, while the 20 \texttt{idiomatic} documents again act as distractors.
Consequently, BM25 is particularly effective at ranking documents containing the same surface form near the top of the ranking, which strongly benefits nDCG@10.

However, BM25 performs substantially worse on R-Precision (36.04), indicating that lexical overlap alone is insufficient to recover the full semantically relevant set.
In particular, BM25 struggles to retrieve \texttt{simplification} and \texttt{sense} documents, where the PIE itself is absent, and retrieval requires semantic abstraction beyond surface-form matching.

The \texttt{span} configuration further highlights this behavior.
Using only the annotated PIE span as the query increases R-Precision from 36.04 to 49.22, but decreases nDCG@10 from 59.07 to 49.87.
This suggests that emphasizing the target expression improves the retrieval of documents that share the same PIE while reducing the need for broader contextual information to obtain highly ranked results.

In contrast, once dense retrieval models are combined with span-focused representations, they consistently surpass BM25.
The span embedding method isolates the semantic span corresponding to the PIE while preserving contextual information, allowing embedding models to focus on localized meaning rather than the full sentence surface form.
As a result, neural models are substantially better equipped to distinguish literal from idiomatic usage and retrieve semantically equivalent paraphrases even when lexical overlap is minimal.

\section{Fine-Tuning Results and Hyperparameters}
\label{apdx:ft_apdx}
We present the full results of the fine-tuning experiments in Table~\ref{tab:ft_full_results}.

We list our fine-tuning hyperparameters in Table~\ref{tab:ft_params}, which are selected based on preliminary experimentation on the validation set.
We use the AdamW optimizer, a \textit{linear} learning rate scheduler, and early stopping.	

With a larger $\alpha_H$, we notice gradient instability that leads to slow convergence.
This could be addressed with additional training data.
When using a smaller $\alpha_H$, the model failed to learn the fine-grained separation of figurative usage that we aim to achieve.

\begin{table}[t]
    \centering
    \small
    \begin{tabular}{lr}
    \toprule
    \textbf{Parameter} & \textbf{BERT-like Models} \\ 
    \midrule
    $\alpha_N$ (``soft'' negatives) & 3 \\
    $\alpha_H$ (``hard'' negatives) & 2 \\
    Max epochs & 10 \\
    Batch size & 32\textasteriskcentered{} \\ 
    Learning rate & 2e-5 \\
    Warm-up steps (linear) & 100 \\
    Min-delta (early stopping) & 0.001 \\
    Patience epochs (early stopping) & 3 \\
    
    \bottomrule
    \end{tabular}
    \caption{Fine-tuning hyperparameters. \textasteriskcentered{} Drama runs with batch size 8.}
    \label{tab:ft_params}
\end{table}

\begin{table}[htb!]
  \centering
  \begin{threeparttable}
  \resizebox{\columnwidth}{!}{
  \begin{tabular}{llll}
    \toprule
    \textbf{Artifact} & \textbf{Type} & \textbf{License} & \textbf{Usage}  \\
    \midrule
    \midrule

    MAGPIE\tnote{a} & Dataset & CC-BY-4.0 & Prompting \\
    Agno\tnote{b} & Framework & Apache-2.0 & Prompting \\
    SentenceTransformer\tnote{c} & Package & Apache-2.0 & Embedding models \\
    PyTorch\tnote{d} & Package & Own & Embedding models \\
    rank-bm25\tnote{e} & Package & Apache-2.0 & BM25 model \\

    \bottomrule
  \end{tabular}
  }
  \begin{tablenotes}
    \small
    \item[a] \url{https://github.com/hslh/magpie-corpus}
    \item[b] \url{https://www.agno.com/}
    \item[c] \url{https://huggingface.co/sentence-transformers}
    \item[d] \url{https://github.com/pytorch/pytorch}
    \item[e] \url{https://pypi.org/project/rank-bm25/}

  \end{tablenotes}
  \end{threeparttable}
  \caption{Artifacts used during experiments, along with their license and usage explanation.}
  \label{tab:artifacts}
\end{table}

\begin{table*}[t]
\centering
\small
\resizebox{\textwidth}{!}{%
\begin{tabular}{llll|cc|cc|cc|cc}
\toprule
&  &  & & \multicolumn{4}{c}{\textbf{w/o instructions}} & \multicolumn{4}{c}{\textbf{w/ instructions}} \\
\cmidrule(lr){5-8}
\cmidrule(lr){9-12}

\textbf{} & \textbf{} & \textbf{} & \textbf{}
& \multicolumn{2}{c}{\textbf{R-Precision}}
& \multicolumn{2}{c}{\textbf{nDCG@10}}
& \multicolumn{2}{c}{\textbf{R-Precision}}
& \multicolumn{2}{c}{\textbf{nDCG@10}} \\
\cmidrule(lr){5-6}
\cmidrule(lr){7-8}
\cmidrule(lr){9-10}
\cmidrule(lr){11-12}

\textbf{Train} & \textbf{Model} & \textbf{Param} & \textbf{I/N}
& \textbf{sentence} & \textbf{span}
& \textbf{sentence} & \textbf{span}
& \textbf{sentence} & \textbf{span}
& \textbf{sentence} & \textbf{span} \\
\midrule

\multirow{5}{*}{\shortstack[l]{\textbf{w/o inst -}\\\textbf{sentence}}}
& SBERT & 110M & - &
42.14\minipm{0.05} & 51.78\minipm{0.10}$^{\scriptsize\blacktriangle}$ &
71.45\minipm{0.07} & 74.60\minipm{0.30}$^{\scriptsize\blacktriangle}$ &
- & - & - & - \\

& Drama$_{\text{1B}}$ & 1B & - &
50.32\minipm{1.39} & 60.73\minipm{1.51}$^{\scriptsize\blacktriangle}$ &
75.48\minipm{1.97} & 80.15\minipm{3.48}$^{\scriptsize\blacktriangle}$ &
- & - & - & - \\

& E5$_{\text{base-v2}}$ & 110M & - &
\textbf{55.41\minipm{0.43}} & \textbf{58.60\minipm{0.30}}$^{\scriptsize\blacktriangle}$ &
\textbf{81.81\minipm{0.32}} & \textbf{85.50\minipm{0.30}}$^{\scriptsize\blacktriangle}$ &
- & - & - & - \\

& BGE-M3 (Dense) & 0.5B & I &
51.37\minipm{0.79} & 55.28\minipm{0.87}$^{\scriptsize\blacktriangle}$ &
77.12\minipm{0.63} & 81.81\minipm{0.63}$^{\scriptsize\blacktriangle}$ &
- & - & - & - \\

& Qwen3-Embedding$_{\text{0.6B}}$ & 0.6B & I &
45.97\minipm{0.75} & 51.32\minipm{0.89}$^{\scriptsize\blacktriangle}$ &
72.02\minipm{0.80} & 74.37\minipm{1.01}$^{\scriptsize\blacktriangle}$ &
- & - & - & - \\

\midrule
\multirow{5}{*}{\shortstack[l]{\textbf{w/ inst -}\\\textbf{sentence}}}

& SBERT & 110M & - &
- & - & - & - &
42.26\minipm{0.08} & 50.51\minipm{0.16}$^{\scriptsize\blacktriangle}$ &
71.53\minipm{0.02} & 71.88\minipm{0.03}$^{\scriptsize\blacktriangle}$ \\

& Drama$_{\text{1B}}$ & 1B & - &
- & - & - & - &
49.23\minipm{1.20} & 59.40\minipm{0.98}$^{\scriptsize\blacktriangle}$ &
74.10\minipm{2.46} & 81.60\minipm{1.50}$^{\scriptsize\blacktriangle}$ \\

& E5$_{\text{base-v2}}$ & 110M & - &
- & - & - & - &
\textbf{54.67\minipm{0.02}} & \textbf{58.12\minipm{0.03}}$^{\scriptsize\blacktriangle}$ &
\textbf{78.68\minipm{0.05}} & \textbf{84.57\minipm{0.04}}$^{\scriptsize\blacktriangle}$ \\

& BGE-M3 (Dense) & 0.5B & I &
- & - & - & - &
46.98\minipm{0.10} & 53.10\minipm{0.06}$^{\scriptsize\blacktriangle}$ &
71.11\minipm{0.14} & 79.63\minipm{0.07}$^{\scriptsize\blacktriangle}$ \\

& Qwen3-Embedding$_{\text{0.6B}}$ & 0.6B & I &
- & - & - & - &
42.14\minipm{1.82} & 42.46\minipm{0.12}$^{\scriptsize\blacktriangle}$ &
64.87\minipm{0.24} & 64.87\minipm{0.23} \\

\midrule
\multirow{5}{*}{\shortstack[l]{\textbf{w/o inst -}\\\textbf{span}}}

& SBERT & 110M & - &
- & 42.00\minipm{0.16} &
- & 69.36\minipm{0.44} &
- & - & - & - \\

& Drama$_{\text{1B}}$ & 1B & - &
- & 49.90\minipm{0.13} &
- & 74.35\minipm{0.15} &
- & - & - & - \\

& E5$_{\text{base-v2}}$ & 110M & - &
- & 55.56\minipm{0.06} &
- & \textbf{84.03\minipm{0.05}} &
- & - & - & - \\

& BGE-M3 (Dense) & 0.5B & I &
- & \textbf{56.36\minipm{0.08}} &
- & 83.33\minipm{0.05} &
- & - & - & - \\

& Qwen3-Embedding$_{\text{0.6B}}$ & 0.6B & I &
- & 49.71\minipm{3.11} &
- & 74.29\minipm{3.51} &
- & - & - & - \\

\midrule
\multirow{5}{*}{\shortstack[l]{\textbf{w/ inst -}\\\textbf{span}}}

& SBERT & 110M & - &
- & - & - & - &
- & 48.03\minipm{0.08} &
- & 71.83\minipm{0.14} \\

& Drama$_{\text{1B}}$ & 1B & - &
- & - & - & - &
- & 52.23\minipm{0.19} &
- & 69.89\minipm{0.08} \\

& E5$_{\text{base-v2}}$ & 110M & - &
- & - & - & - &
- & 51.35\minipm{0.04} &
- & 78.93\minipm{0.04} \\

& BGE-M3 (Dense) & 0.5B & I &
- & - & - & - &
- & \textbf{53.43\minipm{0.07}} &
- & \textbf{79.60\minipm{0.07}} \\

& Qwen3-Embedding$_{\text{0.6B}}$ & 0.6B & I &
- & - & - & - &
- & 49.20\minipm{0.19} &
- & 63.93\minipm{0.06} \\

\bottomrule
\end{tabular}
}%
\caption{Fine-tuning results on the IdioLink dataset averaged over three random seeds.
Values are reported as mean$\pm$std.
I/N indicates whether the model was post-trained with instructions (\S~\ref{ssec:models}); inst denotes instructions.
The best result in each configuration is highlighted in bold.
Symbols ``$^\blacktriangle$'' and ``$^\triangledown$'' indicate an increase or decrease, respectively,
compared to the corresponding sentence embedding setting.}
\label{tab:ft_full_results}
\end{table*}

\section{Artifacts}
\label{apdx:artifacts}

We detail the artifacts we use, their usage, and licenses in Table~\ref{tab:artifacts}.
For the exact versions and more details, see the \textit{requirements.txt} file in the project's repository.

\section{Model Checkpoints}
\label{apdx:checkpoints}

In Table~\ref{tab:checkpoints}, we present the checkpoints (or snapshots) used in this work and the models' sizes and licenses.

\begin{table*}[b]
  \centering
  \small
  \begin{threeparttable}
  \begin{tabular}{llll}
    \toprule
    \textbf{Model} & \textbf{Checkpoint} & \textbf{License} & \textbf{\# Param}  \\
    \midrule
    BGE$_{\text{base-en-v1.5}}$\tnote{a} & BAAI/bge-base-en-v1.5  & MIT & 326M \\
    BGE$_{\text{multilingual-gemma2}}$\tnote{b} & BAAI/bge-multilingual-gemma2 & Gemma & 9B \\
    BGE-M3 (Dense)\tnote{c} & BAAI/bge-m3 & MIT & 0.5B \\
    Contriever\tnote{d} & facebook/contriever & CC-BY-NC 4.0 & 110M \\
    Drama$_{\text{1B}}$\tnote{e} & facebook/drama-1b & CC-BY-NC 4.0 & 1B \\
    E5$_{\text{Mistral}}$\tnote{f} & intfloat/e5-mistral-7b-instruct & MIT & 7B \\
    E5$_{\text{base-v2}}$\tnote{g} & intfloat/e5-base-v2 & MIT & 110M \\
    GritLM$_{\text{7B}}$\tnote{h} & GritLM/GritLM-7B & Apache-2.0 & 7B \\
    GTE$_{\text{Qwen2-1.5B}}$\tnote{i} & Alibaba-NLP/gte-Qwen2-1.5B-instruct & Apache-2.0 & 1.5B \\
    GTE$_{\text{Qwen2-7B}}$\tnote{j} & Alibaba-NLP/gte-Qwen2-7B-instruct & Apache-2.0 & 7B \\
    InstructOR$_{\text{base}}$\tnote{k} & hkunlp/instructor-base & Apache-2.0 & 335M \\
    InstructOR$_{\text{xl}}$\tnote{l} & hkunlp/instructor-xl & Apache-2.0 & 1.5B \\
    Linq-Embed-Mistral\tnote{m} & Linq-AI-Research/Linq-Embed-Mistral & CC-BY-NC 4.0 & 7B \\
    Llama-Embed$_{\text{Nemotron-8B}}$\tnote{n} & nvidia/llama-embed-nemotron-8b & customized-nscl-v1 & 8B \\
    Lychee-embed\tnote{o} & vec-ai/lychee-embed & Apache-2.0 & 1.5B \\
    Multilingual-E5$_{\text{large-instruct}}$\tnote{p} & intfloat/multilingual-e5-large-instruct & MIT & 0.6B \\
    Nomic-v2\tnote{q} & nomic-ai/nomic-embed-text-v2-moe & Apache-2.0 & 0.5B \\
    Qwen3-Embedding$_{\text{0.6B}}$\tnote{r} & Qwen/Qwen3-Embedding-0.6B & Apache-2.0 & 0.6B \\
    Qwen3-Embedding$_{\text{4B}}$\tnote{s} & Qwen/Qwen3-Embedding-4B  & Apache-2.0 & 4B \\
    Qwen3-Embedding$_{\text{8B}}$\tnote{t} & Qwen/Qwen3-Embedding-8B & Apache-2.0 & 8B \\
    SFR-Embedding-Mistral\tnote{u} & Salesforce/SFR-Embedding-Mistral & CC-BY-NC 4.0 & 7B \\
    SBERT\tnote{v} & sentence-transformers/all-MiniLM-L6-v2 & Apache-2.0 & 110M \\
    TART-Contriever\tnote{w} & orionweller/tart-dual-contriever-msmarco & CC-BY-NC 4.0 & 110M \\
    Stella$_{\text{en\_1.5B\_v5}}$\tnote{x} & NovaSearch/stella-en-1.5B-v5 & MIT & 2B \\
    Gemini 2.5 Pro\tnote{y} & gemini-2.5-pro & Proprietary & \textasteriskcentered{}200B+ \\
    Gemini 3 Flash\tnote{z} & gemini-3-flash-preview & Proprietary & \textasteriskcentered{}20B+ \\
    GPT-4o mini\tnote{aa} & gpt-4o-mini-2024-07-18 & Proprietary & \textasteriskcentered{}200B+ \\

    \bottomrule
  \end{tabular}
  \begin{tablenotes}
    \small
    \item[a] \url{https://huggingface.co/BAAI/bge-base-en-v1.5}
    \item[b] \url{https://huggingface.co/BAAI/bge-multilingual-gemma2}
    \item[c] \url{https://huggingface.co/BAAI/bge-m3}
    \item [d] \url{https://huggingface.co/facebook/contriever}
    \item [e] \url{https://huggingface.co/facebook/drama-1b}
    \item [f] \url{https://huggingface.co/intfloat/e5-mistral-7b-instruct}
    \item [g] \url{https://huggingface.co/intfloat/e5-base-v2}
    \item [h] \url{https://huggingface.co/GritLM/GritLM-7B}
    \item[i] \url{https://huggingface.co/Alibaba-NLP/gte-Qwen2-1.5B-instruct}
    \item[j] \url{https://huggingface.co/Alibaba-NLP/gte-Qwen2-7B-instruct}
    \item[k] \url{https://huggingface.co/hkunlp/instructor-base}
    \item[l] \url{https://huggingface.co/hkunlp/instructor-xl}
    \item[m] \url{https://huggingface.co/Linq-AI-Research/Linq-Embed-Mistral}
    \item[n] \url{https://huggingface.co/nvidia/llama-embed-nemotron-8b}
    \item[o] \url{https://huggingface.co/vec-ai/lychee-embed}
    \item[p] \url{https://huggingface.co/intfloat/multilingual-e5-large-instruct}
    \item[q] \url{https://huggingface.co/nomic-ai/nomic-embed-text-v2-moe}
    \item[r] \url{https://huggingface.co/Qwen/Qwen3-Embedding-0.6B}
    \item[s] \url{https://huggingface.co/Qwen/Qwen3-Embedding-4B}
    \item[t] \url{https://huggingface.co/Qwen/Qwen3-Embedding-8B}
    \item[u] \url{https://huggingface.co/Salesforce/SFR-Embedding-Mistral}
    \item[v] \url{https://huggingface.co/sentence-transformers/all-MiniLM-L6-v2}
    \item[w] \url{https://github.com/facebookresearch/tart/tree/main?tab=readme-ov-file#pre-trained-checkpoints}
    \item[x] \url{https://huggingface.co/NovaSearch/stella_en_1.5B_v5}
    \item[y] \url{https://docs.cloud.google.com/vertex-ai/generative-ai/docs/models/gemini/2-5-pro}
    \item[z] \url{https://docs.cloud.google.com/vertex-ai/generative-ai/docs/models/gemini/3-flash}
    \item[aa] \url{https://platform.openai.com/docs/models/gpt-4o-mini}

  \end{tablenotes}
  \end{threeparttable}
  \caption{Checkpoints used during experiments, their license, and their number of parameters. \textasteriskcentered{} = non-official estimation, as this information is not public.}
  \label{tab:checkpoints}
\end{table*}

\section{AI assistants}
\label{apdx:ai_assistants}
We used AI assistants (e.g., ChatGPT) to support code formatting, phrasing suggestions, and LaTeX styling during writing. 
All outputs were reviewed and edited by the authors. 
No content was directly generated or used without human verification.

\end{document}